# SepsisFlow: an event-stream foundation model for continuous risk warning and trajectory modelling in ICU sepsis

Yawen Zhang[1†], Wanzhe Xu[1†], Zhenhua Zhang[1,3], Jinyu Wu[1], Yizhou Zhuang[1], Yitao Yang[2*], Kenta Nakai[1,2*]

[1] *Department of Computational Biology and Medical Science, The University of Tokyo, Kashiwa, Japan*

[2] *Human Genome Center, The Institute of Medical Science, The University of Tokyo, Tokyo, Japan*

[3] *National and Local United Engineering Lab of Druggability and New Drugs Evaluation, Guangdong Provincial Key Laboratory of New Drug Design and Evaluation, School of Pharmaceutical Sciences, Sun Yat-sen University, Guangzhou 510006, China*

** Corresponding authors. E-mail: Yitao Yang (yang-yitao@g.ecc.u-tokyo.ac.jp); Kenta Nakai (knakai@ims.u-tokyo.ac.jp)*

## Abstract

Sepsis is a leading cause of ICU death with heterogeneous onset and course, yet most warning models predict only whether sepsis will occur, leaving clinicians uncertain of its onset context and early trajectory once an alarm sounds. To answer these questions on one hourly clock from ICU admission, we built SepsisFlow: before onset, it combines a Mamba model pretrained on 2.57 million EHRs with ~1,700 hourly features in gradient-boosted trees to predict 6/12/24-h risk; at onset, it fuses Mamba and tabular encoders via a gated network into continuous patient states defining onset contexts; after onset, it encodes 72-h organ-support and physiological trajectories via an attention bidirectional recurrent network into early courses. For pre-onset warning in MIMIC-IV, SepsisFlow detected more cases than NEWS2, qSOFA, and SIRS at equal alarms—gaining ~8 cases per 100 from hourly updating—while outperforming 16 retrained algorithms and maintaining risk ranking in eICU. At onset, its contexts separated 2% cardiovascular-ICU from 15%–17% other 28-day mortality, and within each context after onset, persistent support carried fivefold higher mortality than de-escalation. SepsisFlow establish a foundation-model paradigm for continuous risk and trajectory stratification in critical care.

## Introduction

Sepsis—life-threatening organ dysfunction from a dysregulated host response to infection[1]—caused 166 million cases and 21.4 million deaths in 2021[2], killing over 40% of affected ICU patients[3], a quarter of whom acquire sepsis after admission[4]. Because each hour of delayed antimicrobial therapy increases mortality[5], Sepsis-3's timestamped definition of suspected infection with an acute SOFA rise[1,6] makes pre-onset warning a central goal of critical-care informatics[7]; yet bedside scores discriminate poorly in the ICU[8,9]. To detect deterioration earlier, machine-learning alerts have entered practice: confirming TREWS alerts within 3 h correlated with lower mortality across five hospitals[10], whereas the widely deployed Epic model missed two thirds of cases externally while alerting on 18% of hospitalized patients[11]. Crucially, removing Epic predictions made after clinicians had ordered antimicrobials or cultures reduced its AUROC from 0.62 to 0.47[12], showing that alerts driven by diagnostic or treatment orders merely echo clinical suspicion rather than anticipate pre-treatment deterioration at a manageable alarm burden.

Anticipating deterioration before treatment is difficult because electronic health records (EHRs) entangle patient physiology with clinician actions and the ICU clock. Clinician actions confound the record because test ordering often predicts survival better than test values[13]; moreover, because Sepsis-3 onset is defined from culture and antimicrobial orders[6,7], including diagnostic orders as predictors inflates discrimination[14], just as using measurements before their bedside entry timestamps causes temporal leakage[15]. The ICU clock adds a second confounder: nearly 60% of MIMIC patients have suspected infection at admission[16] and nosocomial risk shifts with length of stay, requiring an explicit clock-prior baseline to isolate physiological signal. Cross-study comparisons face parallel confounders from incompatible onset definitions and scoring units[17]—even after cohort standardization in YAIB[18] and SepsisLab[19], studies still pit single-window bedside scores[8] against continuously updated alerts[11]—so credible early warning must restrict hourly inputs to bedside-visible, non-diagnostic records and benchmark all models and scores under one continuous protocol at matched alarm burden.

Once hourly inputs are restricted to records visible before treatment, risk estimation must rely on how prior clinical events have evolved—the core objective of EHR foundation models. Autoregressive pretraining in CLMBR and Mamba models pretrained on 2.57 million EHRs improved downstream predictions[20] and few-label cross-site adaptation[21], yet their reliance on institution-specific structured codes limits cross-system transfer[22]. This transfer bottleneck is acute in ICU sepsis, where hourly AUROC drops from 0.846 internally to 0.761 externally across international databases[23] and by up to 0.20 in multicentre cohorts[24] due to site-specific care patterns[25], while standard fine-tuning underperforms under distribution shift[26]. Bringing event-stream pretraining into ICU sepsis warning therefore requires pairing portable physiological event sequences with hourly clinical features, verifying that pretraining adds signal beyond architecture alone, and separating transferable risk ranking from local calibration and thresholds[27].

Even when a model warns accurately before onset, binary prediction stops once Sepsis-3 criteria are met, leaving clinicians uncertain of what clinical context onset has occurred in and how the early course will unfold. For onset context, existing phenotypes—such as the α–δ classes[28], SepsisFormer coagulation–inflammation subtypes[29], and dynamic triage clusters[30]— anchor at hospital presentation or triage, which does not coincide with ICU-acquired onset[4] and fails to isolate low-mortality perioperative cardiac-surgery support satisfying Sepsis-3 criteria[31]. For the subsequent course, outcomes diverge across temperature trajectories[32], 72-h SOFA trajectories[33], and organ-dysfunction courses stratifying corticosteroid response[34], yet these trajectory studies start at admission in separate cohorts whereas early-warning studies stop at onset[7,12]. Connecting pre-onset warning to post-onset evolution thus requires a single hourly clock that separates the pre-onset context present at onset from the subsequent 72-h organ-support course evaluated at a fixed landmark[35] with live discharge as a competing risk[36]. To meet these requirements, we introduce SepsisFlow, an event-stream foundation-model framework unifying pre-onset risk warning, onset context phenotyping, and post-onset trajectory modelling on a single hourly clinical clock. Before onset, using only bedside-visible, non-diagnostic records, gradient-boosted trees combine 31-concept encodings from a Mamba model pretrained on 2.57 million EHRs[20] with 1,678 hourly features to predict 6/12/24-h risk; in MIMIC-IV[15], this branch beat 16 retrained algorithms and bedside scores before antimicrobial orders at equal alarm burden, confirmed pretraining and clock-prior gains, and preserved zero-shot risk ranking across 22 eICU hospitals[37]. At onset, a gated network fuses the Mamba and tabular encoders into a continuous patient state resolving three contexts—reproducible by two bedside questions on ICU stay duration and cardiovascular-ICU admission—that separate ~2% post-cardiac-surgery day-28 mortality[31] from 15%–17% mortality in other ICU sepsis. Over the 72 h after onset, an attention recurrent network classifies observed organ-support trajectories under landmark competing-risk evaluation[35,36], revealing fivefold mortality differences between persistent and de-escalating support within the same context while the patient state predicts new 24-h organ failures beyond current SOFA subscores. By linking large-scale EHR pretraining to bedside-verifiable states across the ICU timeline, SepsisFlow advances critical-care decision support from isolated alerts to continuous trajectory stratification.

## Results

### Overview of SepsisFlow

SepsisFlow places sepsis in the ICU on one clinical clock that advances hourly from ICU admission, and answers three questions in turn along this clock (Fig. 1b). Before onset, it estimates at every hour the risk that the patient first meets the Sepsis-3 criteria[1] within 6, 12 and 24 h; at onset, it gives the patient's dynamic state and summarises it as an onset context. After onset, it describes the organ-support course observed in the first 72 h and follows the patient to in-hospital outcome by day 28. The warning before onset and the patient state at onset use only records available at that hour and exclude antimicrobial and culture records, whereas the early course summarizes the records observed after onset (Fig. 1c).

We developed and internally evaluated SepsisFlow in MIMIC-IV[15], excluding stays with sepsis at or before admission, ambiguous admission-year periods or onset <4 h after admission (59% of eligible cases). The cohort comprised 42,028 first ICU stays (9,748 cases, 23.2%), split by patient into development (29,403 stays), validation (8,464) and internal test (4,161; 997 cases) sets (1.9 million hourly scores; Table 1; Supplementary Fig. 1a). In the test set, median onset time was 8.2 h (IQR 5.3–19.9), with 630 cases (63.2%) at 4–11 h; 26.5% of cases versus 16.9% of non-cases entered the CVICU (Table 1). External validation comprised 12,433 stays (421 cases, 3.4%) across 22 eICU-CRD hospitals[37] with microbiology documentation (Table 1; Supplementary Fig. 1a).

SepsisFlow has two branches (Fig. 1c). The risk branch uses two complementary kinds of information. The first is 1,678 hourly clinical features covering vital signs, laboratory tests, organ support, fluid balance, time since ICU admission and the observation process itself (missingness, measurement counts and time since the last measurement). The second is an event stream of numeric events from 31 vital-sign, laboratory, blood-gas and inspired-oxygen concepts, encoded by a Mamba event-stream encoder pretrained on about 2.57 million structured electronic health records[20] and summarised into 32 principal components. The two together make 1,710 inputs, used by gradient-boosted trees to estimate hourly 6-, 12- and 24-h onset risks; the encoder is frozen in this branch. The patient-state branch fuses the same pretrained encoder (last three layers fine-tuned) with the hourly features through a gate to produce a continuous patient state at every scoring hour. The state at the last scoring hour before onset is clustered into onset contexts, the hourly states form state trajectories, and linear models fitted on the frozen state give organ-level risks. The early course is obtained by a separate sequence encoder that clusters the trajectories actually recorded within 72 h of onset, and all test evaluations were performed once per configuration on frozen models.

The following sections follow the clock: when the pre-onset warning is raised, how it compares with published models, what signals it uses, whether it holds in external hospitals, and how onset contexts and 72-h courses unfold.

**SepsisFlow warns hours before onset, mostly before the first antimicrobial order**

On the internal test set, SepsisFlow predicted onset within 24 h with an AUROC of 0.822 (95% CI 0.807–0.836) and an AUPRC of 0.292 (0.270–0.317). For the 12-h and 6-h horizons, the AUROC was 0.843 (0.830–0.856) and 0.859 (0.847–0.871), and the AUPRC was 0.289 (0.264–0.315) and 0.284 (0.258–0.312; Fig. 2a,b; positive prevalence 6.7%, 4.8% and 3.2%). Computed hourly as continuous scores, NEWS2, qSOFA and SIRS reached AUROC 0.525–0.565 and AUPRC 0.075–0.090 at 24 h (Supplementary Fig. 1d).

The alarm threshold was fixed on the validation set (0.187 uncalibrated, 0.191 after Platt calibration; both give the same alarms). At this threshold, 847 of the 997 cases (85.0%) received a first alarm before onset, and 702 (70.4%) had their first alarm 0–24 h before onset, with a median lead of 5.7 h (95% CI 5.4–6.2; Fig. 2c). A warning is useful only if it comes before clinicians start treatment[12]. SepsisFlow uses no antimicrobial or culture inputs, and among these 702 cases the first alarm preceded the first in-ICU antimicrobial order in 555 (79.1%, 76.1–82.1) by a median of 3.0 h (2.6–3.4; Fig. 2c). Of these 702 alarms, 416 (59.3%) came at the first hourly score, and 60.5% (57.5–63.5) of all cases were alarmed 0–24 h before onset and before the first culture or antimicrobial order (Supplementary Table 5). Among cases with onset 4–11 h after admission, 89.8% were alarmed 0–24 h before onset (median lead 5.2 h), compared with 37.1% (median lead 16.5 h, plus 145 alarmed >24 h before onset) when onset occurred ≥12 h after admission (Fig. 2c).

**At the same alarm burden, SepsisFlow identifies more patients about to develop sepsis**

ICU teams often screen for sepsis with qSOFA, SIRS or NEWS2[6,8], and the clinical cost of a warning lies in its alarm burden[38]. We therefore compared first alarms per stay, matching SepsisFlow's validation alarm burden to the published threshold of each score. At the burden of NEWS2 ≥ 5 (46.0 first alarms per 100 patient-days), SepsisFlow identified 76.4% (73.7–79.1) of cases against 56.8% (53.7–59.8) for NEWS2 ≥ 5, 19.7 percentage points more (16.6–22.6). The positive predictive value (PPV) of alarms rose from 16.4% to 21.6%, and alarmed stays per case identified fell from 6.10 to 4.63 (Fig. 2d; Supplementary Table 6). Matched to the lower-burden qSOFA ≥ 2, SIRS ≥ 2 and NEWS2 ≥ 7 thresholds, SepsisFlow identified 38.2, 33.8 and 31.8 percentage points more cases, with PPV higher by 11.2–16.3 percentage points (Supplementary Table 6). When alarms before ICU hour 1 were disallowed, the gain was 21.7 percentage points (18.9–24.5); when all NEWS2 components were required, SepsisFlow still identified 4.7 percentage points more (3.1–6.4; Supplementary Table 7).

Part of this gap reflects the temporal structure of the ICU stay. Cases with onset 4–11 h after admission make up 63%, and SepsisFlow includes time in the unit whereas bedside scores do not. To separate temporal structure from physiological signal, we built a context-only reference model containing only age, sex, first ICU unit and time since ICU admission. At the NEWS2 ≥ 5 burden this reference identified 72.9% (70.0–75.6) of cases, and SepsisFlow identified 3.5 percentage points more (1.9–5.1; Supplementary Fig. 1c). At the lower burdens of SIRS ≥ 2, NEWS2 ≥ 7 and qSOFA ≥ 2, SepsisFlow identified 9.7 (7.3–12.2), 11.7 (9.2–14.4) and 12.8 (10.0–15.8) percentage points more than the reference (Supplementary Table 32). The reference model's 24-h AUPRC was 0.166, compared with 0.287 for the same learner with all hourly features (Table 3) and 0.292 for SepsisFlow (Table 2). SepsisFlow thus sets baseline risk from the ICU clock and uses physiological signal to order alarms.

Continuous hourly re-scoring also outperformed single-point admission screening. At 0.31 first alarms per patient-day, hourly re-scoring identified 70.4% of cases against 62.3% with the first assessment only (+8.1 percentage points, 5.7–10.5), and first-alarm PPV rose from 24.9% to 27.9% (Fig. 2d; Table 3; Supplementary Tables 8–10 and 17).

After calibration on the validation set, SepsisFlow's test calibration slope was 1.01 (0.95–1.08) at 24 h (1.03 at 12 h; 1.01 at 6 h) and its observed-to-expected ratio was 1.03 (0.96–1.11; Supplementary Fig. 1e; Supplementary Table 2). In decision curve analysis at a 10% threshold probability, 24-h net benefit was 0.0240 (0.0206–0.0275) per scoring hour (0.015 at 12 h; 0.009 at 6 h; Fig. 3d; Supplementary Fig. 1f), exceeding NEWS2 (0.0016), qSOFA (0.0004) and SIRS (0) by 0.0224–0.0240 (Supplementary Table 2).

Neither the bedside scores nor the context-only reference was designed for hourly prediction; the stricter comparators are published models developed for the same task.

**Under one protocol, SepsisFlow outperforms 16 published algorithms for the same task**

Because published sepsis models use different cohorts, labels and evaluation rules, we retrained the 16 open-source algorithms from 10 same-task studies with their official code on the same cohort, Sepsis-3 label, split and hourly protocol (Fig. 3; Table 2; Supplementary Table 1). The 16 algorithms comprise signature method + LightGBM[39], EASP XGBoost[40], SeFT-Attn[41], icuDG (Transformer, GRU and TCN)[24], YAIB (Transformer, GRU and TCN)[18],

NCL[42], SepsisLab LSTM[19], the Moor et al. attention model[23], GATSM[43] and Tranchellini et al. (InceptionTime, CNN and LSTM)[26].

On the internal test set, SepsisFlow had the highest AUROC and AUPRC of the 17 models (Fig. 3a,b; Table 2). Against the strongest published algorithm, signature method + LightGBM (AUROC 0.808, AUPRC 0.273), the paired differences were +0.014 (+0.006 to +0.022) and +0.019 (+0.006 to +0.033); against the LSTM of Tranchellini et al. they were +0.075 (+0.064 to +0.087) and +0.125 (+0.107 to +0.144). None of the 32 paired 95% CIs included 0, and all remained significant after Holm adjustment (Fig. 3f; Supplementary Fig. 2a; Supplementary Table 1). The retrained algorithms reached AUROC 0.747–0.808, whereas author weights trained on other databases reached 0.488–0.669 (Supplementary Fig. 2d). At the 6- and 12-h horizons SepsisFlow likewise exceeded all 7 algorithms trained for those horizons, with none of the 14 paired CIs including 0 (Supplementary Fig. 2b,c).

After Platt recalibration fitted on each model's validation set, calibration slopes were 1.01 for SepsisFlow and 0.90–1.17 for the published algorithms (Fig. 3c; Supplementary Table 2). At a 10% threshold probability, SepsisFlow's net benefit (0.0240) exceeded that of 15 of the 16 algorithms (+0.0018 against EASP XGBoost to +0.0079 against Tranchellini LSTM; Fig. 3d; Supplementary Table 2). Between 2 and 48 first alarms per 100 patient-days, SepsisFlow's detection curve lay above every published algorithm, identifying 76.4% of cases against 71.2%–75.9% at the NEWS2 $\geq$ 5 burden (Fig. 3e; Table 2; Supplementary Fig. 2e).

Restricted to the 63,446 scoring hours before the first in-ICU antimicrobial order and culture specimen (4,160 stays, 996 cases), SepsisFlow had an AUROC of 0.834 (0.811–0.855) and an AUPRC of 0.335 (0.306–0.367), again the highest of the 17 models. Its pre-treatment AUPRC exceeded that of all 16 published algorithms (+0.026 against signature method + LightGBM to +0.120 against Tranchellini LSTM; Table 2; Supplementary Table 5).

Table 4 gives the results when unit composition and label rules are varied. When the CVICU was left out (3,362 stays, 733 cases), SepsisFlow had an AUROC of 0.808 (0.791–0.825) and an AUPRC of 0.247 (0.223–0.272), first of the 17 models with all 16 paired CIs excluding 0, and it stayed first when screening swabs were not counted as cultures (808 cases; AUROC 0.810, AUPRC 0.210). Among cases with onset ≥12 h after ICU admission, SepsisFlow's AUROC was 0.759 (0.740–0.778), 0.802 and 0.815 at the 12-h and 24-h landmarks, and 0.773 (0.755–0.790) at stay level with time-matched risk sets, coming ahead in all 20 comparisons against the 16 published algorithms, the context-only reference and the three bedside scores. Outside the CVICU, at the NEWS2 ≥ 5 burden, SepsisFlow identified 72.4% of cases, 9.4 percentage points more than NEWS2 ≥ 5 (6.3–12.7) and 4.8 more than the context-only reference (2.5–6.9). At a low burden of 20 first alarms per 100 patient-days it identified 26.7% of the later-onset cases, more than the three strongest published algorithms and the reference model (5.4 percentage points more than signature method + LightGBM, 1.3–9.8; Table 4; Supplementary Tables 11, 12, 15, 16).

**The pretrained event-stream encoder carries onset information independent of the hourly features**

SepsisFlow's event representation comes from a pretrained foundation model, which raises two questions: whether pretraining gave it information beyond sequence structure alone, and whether it adds to the hourly features. A linear probe on the frozen encoder embedding plus ICU time alone reached an AUROC of 0.765 (0.749–0.781) and an AUPRC of 0.205 (0.189–0.223) at 24 h. An encoder of identical architecture initialised at random reached AUROC 0.692–0.694 and AUPRC 0.159–0.162 across three random seeds, a paired AUPRC difference of +0.043 to +0.045 (Table 3; Supplementary Fig. 3f,g). Pretraining therefore gave the event representation signal related to sepsis onset before it ever saw the labels of this task.

On top of the 1,678 hourly features, the 32 encoder components raised the AUPRC at the 6- and 12-h horizons by 0.013 (0.004–0.023) and 0.010 (0.001–0.018), and on pre-treatment hours by 0.036, 0.024 and 0.016 at 6, 12 and 24 h (+0.012 to +0.018 relative to 32 random-noise columns; Table 3; Supplementary Table 32). Combining the encoder components in a tree model outperformed end-to-end neural fusion (+0.049 AUPRC, 0.036–0.063) and a feed-forward network on the same inputs (+0.060, 0.046–0.075). The full hourly feature set outperformed the 52 YAIB variables with the same context inputs (+0.039 AUPRC, 0.026–0.052), and observation-process features raised the AUPRC by 0.014 (0.006–0.022; Table 3; Supplementary Tables 18 and 32).

**Relative to patients at the same ICU hour, the physiological signals behind the alarms rise towards onset and differ by onset context**

What does SepsisFlow alarm on? We attributed the log-odds of the 24-h output to the inputs with exact interventional TreeSHAP[44] on the validation set (8,464 stays, 1,960 cases; Fig. 4). Each hourly score was explained against 50 development-set scores at the same hour since ICU admission, showing why a patient's risk differs from other patients at the same point of their ICU stay. The inputs were grouped into seven organ concepts (respiratory, metabolism and inflammation, neurology, circulation, kidney, blood and coagulation, liver) plus observation process, static context, fluid therapy, fluid removal, vasopressors and event-stream components (Supplementary Table 33). Under this reference, time since ICU admission takes the same value in a score and in its references, so its direct attribution is zero. Of all 1,710 inputs, the lowest mean arterial pressure (MAP) within 24 h and the most recent Glasgow Coma Scale (GCS) total had the largest mean absolute attributions (Fig. 4a; Supplementary Fig. 3a).

In the 427 validation-set cases with scores both >24 h and in the last 6 h before onset, the within-patient total contribution of the seven organ concepts rose from 0.493 to 1.017 log-odds (+0.524, 0.456–0.596; higher in 322 of 427), led by respiratory (+0.163), metabolism and inflammation (+0.153) and neurological signals (+0.117; Fig. 4b,c). In five matched validation-set patients without sepsis per case, the contribution of physiological signals did not rise (change −0.111, −0.142 to −0.075), giving a case–control contrast of +0.635 log-odds (0.562–0.710; Fig. 4c; Supplementary Table 34). In the last 6 h before onset across all 1,960 cases, the seven organ concepts contributed more than the observation process (0.607 against 0.427 log-odds; Supplementary Fig. 3b,c; Supplementary Tables 30 and 34).

The contribution of the lowest MAP was highest just above 70 mmHg, the cardiovascular SOFA cut-off of the onset definition: a patient still without onset whose lowest MAP is just above 70 mmHg can be one cardiovascular SOFA point from meeting the definition. $PaO_2/FiO_2$ and platelet count showed the same steps at their SOFA cut-offs (about 300 mmHg and 150 × $10^9$/L), and the contribution of creatinine fell above 1.8 mg/dL, near its 2.0 mg/dL cut-off (Fig. 4a; Supplementary Fig. 3e; Supplementary Table 35).

Among the 1,581 validation-set cases with a frozen onset context, respiratory signals accounted for 47.4% (44.7–50.2) of positive physiological attribution in Early onset, CVICU cases, against 21.4% and 17.2% in Late ICU onset and Early onset, non-CVICU cases (Fig. 4d). Neurological signals were highest in Late ICU onset (20.2%), whereas circulation (16.4%), kidney (15.1%) and blood and coagulation signals (14.9%) were highest in Early onset, non-CVICU cases (single-patient example in Supplementary Fig. 6).

**Without retraining, SepsisFlow still ranks onset risk in external hospitals**

Differences in measurement frequency, documentation and care practice between ICU databases are common, and discrimination often falls when a model is transferred to a new hospital[23,24,26]. We applied SepsisFlow unchanged, with frozen weights and no local training, to the 22 eICU-CRD hospitals[37] with microbiology records (421 cases defined by antimicrobial use among 12,433 stays). SepsisFlow predicted onset within 6, 12 and 24 h with

AUROCs of 0.738 (0.714–0.762), 0.724 (0.699–0.750) and 0.688 (0.659–0.717); the AUROC was 0.714 in the hospital contributing most cases and 0.717 in the other 21 pooled (Supplementary Fig. 5a,c; Supplementary Table 23).

What has to be reset locally after transfer is the probability scale and the alarm threshold. After 5-fold cross-fitted local logistic recalibration, the observed-to-expected ratio in the external cohort was 0.99 (0.89–1.11) with a calibration slope of 0.97 (Supplementary Fig. 5b; Supplementary Tables 23 and 26).

The external test concerns the hourly warning before onset. At the moment of onset, the next bedside question is what state the patient is in.

**The onset patient state clusters into three contexts reproducible at the bedside**

Onsets that equally meet the Sepsis-3 criteria can occur in very different clinical contexts. Published sepsis phenotypes such as the α–δ phenotypes of Seymour et al.[28] and the coagulation–inflammation subphenotypes of SepsisFormer[29] require a full set of laboratory results that is not always available at onset. SepsisFlow's patient-state branch provides a continuous state representation at every scoring hour. Clustering the state at the last scoring hour before onset (fitted and frozen on the discovery set) divided the 997 test-set cases into three contexts (Fig. 5a):

- Late ICU onset (359 cases): onset ≥ 12 h after ICU admission;
- Early onset, non-CVICU (407 cases): onset 4–12 h after ICU admission outside the CVICU;
- Early onset, CVICU (231 cases): onset 4–12 h after ICU admission in the CVICU.

These three contexts can be reproduced by two bedside questions: did onset occur within 12 h of ICU admission? If so, was the first ICU the CVICU? The two-question rule agreed with the contexts in 87.9% of test-set cases (85.8–89.8; Cohen's κ = 0.81) and in 83.0%–88.7% of the discovery, tuning and replication sets (Fig. 5b; Supplementary Table 3). A feed-forward network trained on the hourly features, whose hidden states were clustered with the same frozen procedure, reproduced the three contexts (adjusted Rand index [ARI] 0.795, 0.752–0.838; Supplementary Table 25), so this structure does not depend on one particular encoder.

The three contexts differ in their organ-dysfunction profile (Fig. 5c; Supplementary Table 19). A respiratory SOFA component ≥ 2 at onset was present in 43.3% of the Early onset, CVICU group against 25.6% and 13.5% of the Late ICU onset and Early onset, non-CVICU groups; a central nervous system (CNS) component ≥ 2 was most frequent in Late ICU onset (20.9%), and a liver component ≥ 2 occurred in <5 Early onset, CVICU patients. With live discharge as a competing event, cumulative 28-day in-hospital death was 17.1% (13.4–21.2), 15.1% (11.8–18.7) and 2.2% (0.8–4.8) in the Late ICU onset, Early onset, non-CVICU and Early onset, CVICU groups (Gray's test $P = 2.4 \times 10^{-7}$; Fig. 5d), and crude mortality in the two-question rule groups was 16.2%, 15.7% and 0.9% (Fig. 5b). The Early onset, CVICU profile matches the known situation in which transient organ support after cardiac surgery with prophylactic antimicrobials meets the Sepsis-3 criteria[31] (Supplementary Fig. 5d; Supplementary Tables 15, 20 and 23).

The onset context describes the axis of when and where onset occurs and complements laboratory-based phenotypes (Fig. 5e; Supplementary Fig. 4c). Of the 231 Early onset, CVICU cases, 201 (87%) were Seymour α[28]; Late ICU onset was mostly γ (54%); Early onset, non-CVICU split evenly between β and γ. The adjusted Rand index between the contexts and the Seymour phenotypes was 0.18 (0.15–0.21), and with the coagulation–inflammation subphenotypes[29] −0.002 (−0.024 to 0.029). The two-question rule assigned almost every case (995 of 997), whereas the coagulation–inflammation subphenotypes, which need a complete laboratory profile, assigned only 47.3% of cases without imputation (Supplementary Table 25). In the pre-registered prognostic comparison, both the onset contexts and the two-question rule improved the discrimination of 28-day in-hospital death beyond the basic adjustment (Wolbers C-index[45] increments of +0.061 and +0.070), and the prognostic information of the contexts is almost entirely carried by the two-question rule (Supplementary Table 25).

The three contexts summarise the patient state for bedside use, and the continuous state itself carries prognostic information beyond bedside variables. Adding the onset patient state to a bedside model of the two-question rule, age, SOFA score at onset, vasopressor use and invasive ventilation raised the AUROC for a persistent organ-support course from 0.860 (0.832–0.883) to 0.896 (0.873–0.917; +0.036) and for 28-day in-hospital death from 0.786 (0.742–0.827) to 0.835 (0.797–0.868; +0.049; Supplementary Table 21).

The onset context sets the baseline risk at onset, but outcomes within the same context still differ widely. Moving on along the clock, we follow the first 72 h after onset.

**Within the same onset context, the organ-support course in the 72 h after onset leads to different outcomes**

Previous studies have mostly described the course after onset by trajectories of the SOFA score[33] or temperature and vital signs[32]. SepsisFlow encodes the organ-support and physiological trajectories recorded within 72 h of onset, yielding three observed early courses among 995 test-set cases (Fig. 6): de-escalating support (404 cases), early ICU discharge (260 cases) and persistent organ support (331 cases). At 24 h after onset, 86.6% of the persistent organ-support group remained on advanced respiratory support against 4.6% of the de-escalating group, and trajectories of noradrenaline-equivalent dose, renal replacement therapy and GCS diverged as well (Fig. 6d; Supplementary Fig. 4e). Of the 225 Early onset, CVICU patients still in hospital at 72 h, 204 (90.7%) followed early ICU discharge, whereas Late ICU onset and Early onset, non-CVICU patients split between de-escalation and persistent support (Fig. 6a).

The divergence of courses corresponds to a divergence of outcomes. From the 72-h landmark (891 patients alive in hospital; 84 in-hospital deaths before day 28), day-28 in-hospital mortality was 21.5% (17.3–26.6), 4.4% (2.7–7.2) and 2.0% (0.8–4.5) for persistent organ support, de-escalation and early ICU discharge (Fig. 6b). The difference remained within the same onset context: 24.3% against 4.7% in Late ICU onset and 20.4% against 4.3% in Early onset, non-CVICU (23.2% against 4.3% outside the CVICU; Fig. 6b; Supplementary Table 15). In cause-specific Cox models with live discharge as a competing event, the hazard ratio for persistent support against de-escalation was 3.12 (1.77–5.49) unadjusted, 4.10 (2.15–7.81) after adjustment for age, sex, onset SOFA, vasopressors, invasive ventilation and onset context, and 2.80 (1.42–5.55) after further adjustment for SOFA at 72 h (Fig. 6c; Supplementary Tables 27–29).

Added to the onset context, the SepsisFlow early course raised Harrell's C-index by 0.064 (0.020–0.107) and the Wolbers C-index by 0.098 (0.062–0.133, from 0.639 to 0.737), exceeding the SOFA-trajectory subtypes of Xu et al., the temperature-trajectory subtypes of Bhavani et al. and latent-class organ-support trajectories (+0.046 and +0.047 at most; Supplementary Fig. 4f; Supplementary Tables 4, 22 and 31).

The patient state joins the 6 h before onset and the 72 h after onset into one trajectory (Fig. 6f,g). A linear model fitted on the discovery set discriminated the persistent organ-support course with an AUROC rising from 0.795 (0.737–0.846) at 24–18 h before onset to 0.896 (0.874–0.916) at 6–0 h before onset, and from 0.795 to 0.863 in the 278 patients observed for ≥24 h before onset (Fig. 6g). At 6–0 h before onset, the patient state exceeded a bedside model of current organ support, SOFA components, onset context and ICU time (0.896 against 0.860; +0.037, +0.015 to +0.057; Supplementary Table 21).

Finally, the same patient state indicates the organ that will be affected next (Fig. 6e). At hourly scoring points 0–72 h after onset, a linear model predicting a new SOFA component ≥ 2 within the next 24 h discriminated better from the patient state than from the six current SOFA components for the respiratory (AUROC difference +0.193, Holm-adjusted 95% CI

+0.138 to +0.248), cardiovascular (+0.143, +0.056 to +0.232), central nervous (+0.131, +0.076 to +0.189) and renal systems (+0.130, +0.056 to +0.201). At scoring points with at least one organ about to fail within 24 h, the top-ranked organ from the patient state was correct in 63.3% (60.1–66.5) against 54.4% (50.6–58.3) for the current SOFA components (Supplementary Table 21).

Along one ICU clock, SepsisFlow gives a chain of judgements that unfolds in time from hourly pre-onset warning to onset context, 24-h next-organ risk and 72-h organ-support courses.

## Discussion

SepsisFlow places the early warning of ICU sepsis, the patient state at onset and the early course after onset on one clinical clock that advances hourly from ICU admission, in a structure of one clock and two branches. The risk branch gives the risk of onset within 6, 12 and 24 h at every hour; the patient-state branch gives a continuous patient state, summarises it at onset as an onset context that two bedside questions reproduce, and uses it after onset to indicate the organ likely to fail next; a separate sequence encoder classifies the 72-h organ-support course. Earlier studies examined warning, subtyping and course in different cohorts and from different time origins[7,28,33]. In SepsisFlow, the warning and the patient state use only records visible before their own time point, and the three judgements connect along one clock.

The credibility of the warning rests first on the temporal semantics of its inputs. SepsisFlow takes the visibility time of every record as the later of its measurement time and its entry time, and excludes from the inputs the antimicrobial orders and culture specimens that make up the onset label. This reduces on the input side the problem identified by Kamran et al., that a model learns the care clinicians have already begun[12]; on the scoring hours before the first in-ICU antimicrobial order and culture, SepsisFlow's AUROC and AUPRC were still the highest of the 17 models. It rests second on the comparison protocol: because published metrics cannot be compared directly across different cohorts and labels[17,18], 16 published algorithms were retrained with their official code on the same cohort, label, split and hourly protocol. SepsisFlow's AUROC and AUPRC remained the highest when the CVICU was excluded, when the culture definition was changed, and in later-onset and landmark analyses.

The contribution of the pretrained event-stream encoder should be considered separately from that of the hourly features. A linear model on the frozen encoder representation plus ICU time discriminated clearly better than an encoder of identical architecture initialised at random, which shows that pretraining gave the representation information related to onset before it ever saw the labels of this task. On top of almost 1,700 hourly features, this representation added measurably at the 6- and 12-h horizons and on the scoring hours before treatment, and did so relative to the same number of random-noise columns. Combining the encoder components in gradient-boosted trees outperformed neural-network fusion of the same frozen encoder (AUPRC +0.035), end-to-end fusion (+0.049) and a feed-forward network trained on the same inputs (+0.060). The value of foundation models for electronic health records has been questioned because they are rarely evaluated on tasks of clinical importance[20,21]; in this study, the pretrained representation added measurably when it entered a tabular learner that includes the observation process and ICU time. At the same time, the context-only reference shows that time since ICU admission and the first ICU unit are a strong prior, and the increment of SepsisFlow over this reference grew from 3.5 to 12.8 percentage points as the alarm burden tightened. Relative to other patients at the same ICU hour, the contributions of respiratory, metabolism and inflammation and neurological signals rose within the same patients as onset approached, alongside the observation process.

The clinical cost of a warning lies in the alarm burden it creates[38], so all alarm-level comparisons in this study were made at the same burden with the first alarm per stay as the unit. Under this condition, hourly re-scoring of SepsisFlow identified 8.1 percentage points more cases than alarming at the first assessment only. Deployed models are updated continuously[11], whereas bedside scores are usually computed once in evaluations[8], so comparisons of warning tools should report the monitoring mode separately from the model itself. Transferred without retraining to 22 eICU hospitals, SepsisFlow's 24-h AUROC was 0.688, third of the 17 models behind two hand-crafted feature tree models, while the probability scale and alarm threshold had to be reset locally[25,27]. Tranchellini et al. found that cross-hospital deployment depends on the data conditions of the target hospital[26], and the effect of documentation differences between databases on event-stream representations warrants dedicated study.

That the patient state at onset carries prognostic information beyond bedside variables has to do with its origin. Produced by gated fusion of the fine-tuned encoder with the hourly features under a competing-risk objective, it is exposed to the evolution of the hourly inputs and event history rather than only the current cross-section. That the three onset contexts can be reproduced by two questions—whether onset occurred within 12 h of ICU admission and whether the first ICU was the CVICU—shows that the main structure of the onset state is when and where onset occurs. Published sepsis phenotypes are mostly defined by laboratory profiles at presentation[28,29], which could be assigned to 47.3% of cases without imputation, whereas the two-question rule assigned 995 of 997 cases. The profile of the Early onset, CVICU context (2.2% 28-day mortality, predominantly respiratory dysfunction and 87% Seymour α) matches the situation reported by Howitt et al., in which transient organ support after cardiac surgery with prophylactic antimicrobials is enough to meet the Sepsis-3 criteria[31]. SepsisFlow marks these onsets out so that they can be considered separately in alarming and stratification.

After onset, the 72-h course actually observed separated day-28 in-hospital mortality about fivefold within the Late ICU onset and Early onset, non-CVICU contexts. The hazard of death for persistent organ support relative to de-escalation was independent of age, sex, severity, organ support and context at onset and went beyond what SOFA at 72 h explains. Landmark analysis and the handling of competing risks are necessary here: they include only patients still in hospital at 72 h[35] and treat live discharge as a competing event[36] rather than as uninformative censoring, without which the low mortality of early ICU discharge would be counted wrongly. Compared with SOFA trajectories[33], temperature trajectories[32] and latent-class organ-support trajectories, the SepsisFlow early course added a larger Wolbers $\Delta C$ on top of the onset context, with paired differences excluding 0. The patient state can be evaluated before and after onset: a persistent organ-support course could already be distinguished by the state 6 h before onset better than a bedside model of current organ support, SOFA components, context and ICU time. After onset, the same state indicated, better than the current SOFA components, the organ that will newly fail within 24 h for the respiratory, cardiovascular, central nervous and renal systems.

Several points should be made clear. SepsisFlow was developed in a single health system with a random split by patient, and external validation covered only the warning before onset using antimicrobials to define infection, with cases concentrated in one eICU hospital. SepsisFlow does not forecast the multi-organ course: its organ-level models cover the next 24 h only, its three early courses classify the records observed in the 72 h after onset rather than predicting them, and pre-onset risk did not predict the later course beyond a bedside onset model. The event-stream encoder processes numeric events of 31 concepts without treatments, antimicrobials, cultures or time gaps between events, and the model gives no estimate of its own uncertainty. Treatment-effect heterogeneity by onset context or early course could not be estimated with these data, and the explanation analyses show how the model uses its inputs rather than causal effects.

Testing this chain of judgements in practice will require local recalibration and alarm-threshold setting on local data first, followed by evaluation within the clinical alarm workflow to see whether lead time leads to earlier assessment and treatment. When the warning before onset, the patient state at onset and the early course after onset are placed on one clinical clock using only records visible at the time, computational support for ICU sepsis can move from an isolated alarm towards judgements that unfold along the course and can be checked step by step at the bedside.

## Methods

### Study design, data sources and ethics

This is a retrospective model development and validation study. The model was developed and evaluated internally on MIMIC-IV (version 3.1)[15] and validated externally on the eICU-CRD (version 2.0)[37], both accessed through PhysioNet credentialed data use agreements. MIMIC-IV was approved by the institutional review boards of Beth Israel Deaconess Medical Center and the Massachusetts Institute of Technology with a waiver of informed consent, and eICU-CRD was certified as meeting safe-harbour standards. Along one clinical clock advancing hourly from ICU admission, SepsisFlow estimates pre-onset sepsis risk and the onset patient state using only records visible before each time point, and classifies the trajectories recorded in the 72 h after onset.

### Cohort and data split

Starting from 65,354 adult first ICU stays in MIMIC-IV, we excluded stays meeting onset criteria at or before admission (163), stays with ambiguous admission-year periods (9,162) and stays with onset <4 h after admission (14,001; 59% of eligible cases); onset at 4 h and short non-sepsis stays were kept. The cohort of 42,028 stays (9,748 cases, 23.2%) was split at random by patient into development (29,403 stays, 6,791 cases, 1,339,337 scores), validation (8,464; 1,960; 379,273) and internal test (4,161; 997; 182,415) sets (Supplementary Fig. 1a;

Table 1). The development set was used for model fitting, the validation set for early stopping, model selection, thresholds and calibrators, and the internal test set once per configuration for evaluation. The validation set was further split by patient into alarm-development (20%, 379 cases), discovery (50%, 984 cases), tuning (15%, 295 cases) and replication (15%, 302 cases) subsets for threshold matching and subtyping. Scoring points are every whole hour after ICU admission up to the earliest of onset, ICU death or ICU discharge.

**Operational definition of sepsis onset**

Onset was operationalised in the electronic record according to the Sepsis-3 criteria[1,6], following the sepsis definition of the MIMIC code repository[16]. Suspected infection was determined by pairing a microbiology specimen with the start of an in-ICU antimicrobial order: an order within 72 h after a specimen, or a specimen within 24 h after an order; the time of suspected infection is the earlier of the two (screening swabs are excluded in Supplementary Table 16). Organ dysfunction was a SOFA score of at least 2 over a rolling 24-h window ending between 48 h before and 24 h after suspected infection, with missing components scored 0. The onset time $T_0$ is the later of suspected infection and SOFA ≥ 2 for the first qualifying pair.

**Visibility time of records**

The measurement time of a record is not the time at which it becomes visible at the bedside. We defined the visibility time of every record as the later of its measurement time and its entry time, so that the set of records available at prediction time t is

$$\mathcal{V}(t) = \{r : t_{\text{event}}(r) < t, \quad \max\{t_{\text{event}}(r), t_{\text{store}}(r)\} < t\}$$

where t_event is the measurement time and t_store its entry or laboratory result time (or the later of event and entry offsets in eICU-CRD). All hourly features, event streams and bedside scores were built under this rule, and antimicrobial orders and culture specimens enter no input. An audit after model development found that 19 respiratory-support features took the start of a ventilation interval as their visibility time; these 19 columns were removed from the risk branch, which was refitted, and were recomputed with repaired timestamps when patient states were evaluated.

**Hourly clinical features**

The tabular input at each scoring hour is a frozen frame of 1,696 columns covering vital signs, laboratory tests, blood gases, GCS, oxygen therapy and ventilator settings, urine output, renal replacement therapy, fluid balance, blood products, vasopressors and static context variables. Continuous measurements are represented over 6-h and 24-h windows by most recent value, mean, minimum, maximum, standard deviation and change; treatments by status and delivered amounts. Variables that can be missing carry observation-process features: missingness indicator, measurement count, window observation flag and time since last measurement (926 columns after removing the 19 columns with future timestamps). Missing values are imputed with the development-set median, and continuous variables are clipped at the 1st and 99th percentiles and standardised using frozen development-set statistics. After removing the 19 columns, 1,677 columns remained, and the ICU-time channel

$$\varphi(t) = \frac{\ln(1 + \min(t, 72))}{\ln 73}$$

(t being the hours since ICU admission) was added, giving 1,678 hourly features.

**Event-stream encoder and its representation**

The event stream consists of numeric events of 31 concepts (seven vital signs, 18 laboratory and coagulation tests, and six blood-gas and $FiO_2$ variables) and contains no treatments, antimicrobials, cultures or time gaps between events. The most recent up to 4,093 events visible before the scoring hour are mapped to tokens by the numeric bins of the pretrained tokenizer and passed to the Mamba state-space model pretrained by Wornow et al. on 2.57 million patients[20,46] (24 layers, hidden size 768, 121 million parameters). The final-layer hidden state at the end symbol is taken as the 768-dimensional event representation e(t). In the risk branch the encoder is frozen, and the event representation is compressed by principal component analysis, fitted on a stratified 20% sample of development-set stays, into 32 components (84.3% of the variance):

$$z(t) = W^{\top}(e(t) - \mu), \; e(t) \in \mathbb{R}^{768}, \; z(t) \in \mathbb{R}^{32}$$

where μ is the sample mean and W holds the loadings of the first 32 components.

**Risk branch**

The risk branch combines the 1,678 hourly features and the 32 encoder components (1,710 inputs in all) in gradient-boosted trees[47], fitting one binary classifier for first onset within 6, 12 and 24 h:

$$\hat{p}_H(t) = f_H\big(x(t), \varphi(t), z(t)\big), \; H \in \{6, 12, 24\}$$

The models were fitted on the development set with early stopping on the validation-set AUPRC, and tree hyperparameters were chosen on the validation set among six pre-specified configurations. The output of each horizon is calibrated by Platt scaling fitted on the alarm-development subset of the validation set for calibration analyses and decision curves. The alarm rule raises one alarm per stay at the first scoring hour at which the uncalibrated 24-h output reaches the threshold θ:

$$A = \min\{t : \hat{p}_{24}(t) \geq \theta\}, \; B(\theta) = 100 \cdot \frac{n_{\mathrm{alarm}}(\theta)}{D}$$

The threshold θ was fixed on the alarm-development subset of the validation set to a pre-specified alarm burden of 0.31 first alarms per patient-day (0.187 uncalibrated, equal to 0.191 after calibration). Detection was defined as a first alarm 0–24 h before onset, and alarm burden as the number of first alarms per 100 patient-days at risk. In the comparison of hourly re-scoring with a single assessment, the single assessment could alarm only at the first scoring hour after ICU admission at the same validation burden. The context-only reference model uses the same learner with age, sex, first ICU unit and the ICU-time channel as its only inputs.

**Patient-state branch**

The patient-state branch shares the pretrained encoder with the risk branch and is trained on its own (Fig. 1c). The 1,696 hourly columns go through a tabular network to $h_{tab}(t)$, the event vector and ICU time go to $h_{es}(t)$, and a gate g(t) combines them:

$$h(t) = g(t) \odot h_{\mathrm{es}}(t) + \left(1 - g(t)\right) \odot h_{\mathrm{tab}}(t), g(t) = \sigma(\mathrm{MLP}[h_{\mathrm{tab}}(t), h_{\mathrm{es}}(t)])$$

where σ is the sigmoid function and ⊙ the element-wise product. The patient state feeds a discrete-time competing-risk head[48]: the 72 h after the scoring hour are divided into 72 one-hour intervals, and in each interval a softmax gives the conditional probabilities of no event, sepsis onset and ICU death, from which the overall survival and cause-specific cumulative incidence functions follow:

$$S(\tau) = \prod_{s \le \tau} P(0 \mid s, h), F_k(\tau) = \sum_{s \le \tau} P(k \mid s, h)\, S(s - 1)$$

The training objective is the competing-risk negative log-likelihood, supplemented by a binary loss for onset within 6, 12 and 24 h and a squared penalty that draws each gate element towards a preset target:

$$\mathcal{L} = -\sum_{i} [\mathbb{1}(\delta_i > 0)(\log P(\delta_i \mid \tau_i, h_i) + \log S(\tau_i - 1)) + \mathbb{1}(\delta_i = 0) \log S(\tau_i)] + \lambda_1 \mathcal{L}_{\mathrm{BCE}} + \frac{\lambda_2}{H} \sum_{j=1}^{H} \left(g_j - g^*\right)^2$$

where $\delta_i$ and $\tau_i$ are the event type (0 for censoring) and interval of sample i, H = 256, $\lambda_1$ and $\lambda_2$ are weights and g* is the gate target. Training proceeded in three phases—the first two with the encoder frozen and the third fine-tuning its last three layers—and the model was selected by validation-set 24-h AUPRC and frozen. The patient state does not enter the risk branch or the alarm (Supplementary Table 18), and post-onset states are described only.

**Onset contexts**

Onset contexts are clusters of the patient state at the last scoring hour before onset, h($T_0$−), fitted and frozen on the discovery set. Standardised by discovery-set statistics, the state goes through a projection network to a 64-dimensional z and K centres $\mu_j$ with a Student's t kernel under structure-preserving deep embedded clustering[49]:

$$q_{ij} = \frac{\left(1 + \|z_i - \mu_j\|^2\right)^{-1}}{\sum_{j'}\left(1 + \|z_i - \mu_{j'}\|^2\right)^{-1}}, \; p_{ij} = \frac{\frac{q_{ij}^2}{\sum_i q_{ij}}}{\sum_{j'}\frac{q_{ij'}^2}{\sum_i q_{ij'}}}, \mathcal{L}_{\text{ctx}}$$
$$= \text{KL}(P \parallel Q) + \beta \sum_i \|\hat{h}_i - h_i\|^2 + \gamma\, \text{KL}(\bar{q} \parallel u)$$

where $\hat{h}_i$ is the state reconstructed from $z_i$, $\bar{q}$ is the mean soft assignment over samples, u the uniform distribution and β and γ weights. Candidate values of K were 3 to 6, each fitted on the discovery set with five random seeds and assessed on the tuning set without outcome data; K = 3 was chosen for having the highest cross-seed stability (mean pairwise ARI 0.850 in the discovery set and 0.821 in the tuning set). The final partition is a consensus of the five seeds via average-linkage hierarchical clustering on co-association distances, named Late ICU onset, Early onset, non-CVICU and Early onset, CVICU. The two-question bedside rule (onset <12 h of ICU admission; first ICU = CVICU) and the FastICA projection were fixed on the discovery set and evaluated on the tuning, replication and test sets (Supplementary Table 3).

**Early courses**

Early courses are clusters of the post-onset trajectories from $T_0$ to $T_0$ + 72 h (or earlier ICU discharge or death), fitted on the discovery set without outcome data (Fig. 6). Standardised across 258 channels of 55 concepts (values, observation flags and organ-support doses), each trajectory goes through a two-layer bidirectional LSTM (128 units each way) with attention pooling to a 256-dimensional vector and a reconstruction decoder:

$$\mathcal{L}_{\text{course}} = \frac{\sum_{t,d} m_{td}(\hat{x}_{td} - x_{td})^2}{\sum_{t,d} m_{td}} + \beta\,\mathrm{KL}(P \parallel Q) + \gamma\,\mathrm{KL}(\bar{q} \parallel u) - \eta\,\mathrm{H}(\bar{q})$$

where $m_{td}$ indicates whether channel d was observed at hour t, H is the entropy and η a weight. Candidate values of K = 3 to 6 were fitted with five random seeds on the discovery set and assessed on the tuning set; only the pre-specified target K = 3 met the cross-seed stability rule (median pairwise ARI ≥ 0.75, minimum ≥ 0.50, and minimum class-wise median Jaccard ≥ 0.60). The frozen model assigned the 995 test-set cases with a valid trajectory into De-escalating, Early ICU discharge and Persistent organ support.

**Classifiers on the patient state**

The discrimination of the later course by the pre-onset state was evaluated in 6-h bins from 24 h before to 24 h after onset: an L2-regularised logistic regression (regularisation chosen by 5-fold cross-validation on the discovery set) was fitted on the discovery-set states of each bin to separate persistent organ support from the other two courses, both overall and in the 278 patients observed for ≥24 h before onset. The bedside comparator for the same bins uses current vasopressor use, advanced respiratory support, renal replacement therapy, the six SOFA components, onset context and log ICU time. The prognostic increment of the onset state beyond a bedside model (two-question rule, age, SOFA at onset, vasopressor use and invasive ventilation) was evaluated on the test set by adding the out-of-fold discovery-set state score for persistent organ support and 28-day in-hospital death (Holm-adjusted paired bootstrap intervals). For organ-level risk at hourly points 0–72 h after onset, an L2 logistic regression on the development-set patient state and hours since onset predicted whether each organ with current SOFA < 2 would reach ≥ 2 within 24 h, compared on the test set against the same model on the six current SOFA components with Holm adjustment.

**Bedside scores and matching of alarm burden**

NEWS2, qSOFA[6] and SIRS were computed at every scoring hour under the same visibility rule, each component taking the most recent value within the 24 h before the scoring hour and missing components scoring 0. As continuous scores they were evaluated by AUROC and AUPRC; as alarm rules, the first hour per stay at which NEWS2 $\geq$ 5, NEWS2 $\geq$ 7, qSOFA $\geq$ 2 or SIRS $\geq$ 2 was reached was taken. For the comparison with each rule, SepsisFlow's threshold was chosen on the alarm-development subset of the validation set as the threshold whose alarm burden was closest to, without exceeding, that of the rule, and frozen for the test set (sensitivity analyses in Supplementary Tables 7 and 9).

**Comparison with published algorithms under one protocol**

We searched for adult ICU Sepsis-3 models published from 2019 to 2026 in first-quartile journals or CCF-A conferences that predict before onset from structured records alone (Supplementary Table 1). The 16 algorithms from 10 studies were retrained with their official code (NCL as a PyTorch port of the official code) on the same cohort, labels, split and hourly protocol: signature method + LightGBM with its 40 raw variables[39]; EASP XGBoost with its 160 features[40]; SepsisLab LSTM with tokenised observation records of its 21 variables[19]; and the other 13 algorithms with the 52 YAIB variables[18] mapped onto this study's hourly frame plus the ICU-time channel (65 columns). Author-released weights were applied by direct inference where the licence allowed: weights trained on other databases were used on the internal test set, and weights trained on MIMIC-IV only on eICU-CRD; all 81 scored configurations had their alarm threshold and Platt calibrator set on their own validation predictions only (Supplementary Table 1).

**Ablations and component contributions**

Ablations were performed on frozen models and predictions, each evaluated once on the internal test set (Table 3; Supplementary Table 18). They comprised removal of the 32 encoder components; replacement of the encoder components by 32 columns of standard normal noise; comparison of the frozen encoder with an identical architecture initialised at random (three seeds) under a linear probe on the representation plus ICU time; end-to-end neural-network fusion and a feed-forward network on the same inputs; progressive input tiers; switching off the observation-process features; and Platt calibration against the raw output.

**Model explanation**

The 24-h risk output was attributed to the 1,710 inputs with the exact interventional algorithm of TreeSHAP[44] on the log-odds scale on the validation set. The reference for each scoring hour was 50 development-set scores at the same whole hour since ICU admission, so ICU time takes the same value in a score and in its references and has zero attribution. The inputs were grouped into seven organ concepts plus ICU time, observation process, static context, fluid therapy, fluid removal, vasopressors and event-stream components under a fixed clinical-system mapping (Supplementary Table 33). Within-patient attribution changes from >24 h to the last 6 h before onset were evaluated in 427 validation cases against five matched validation-set patients without sepsis per case using 2,000 matched-set bootstrap resamples (Supplementary Table 34).

**External validation**

External validation used the 12,433 adult first ICU stays (421 sepsis cases, 738,424 hourly scoring points) of the 22 eICU-CRD hospitals with microbiology records after the same early-onset exclusions. Culture and antimicrobial records rarely co-occur in the same hospital in these data, so suspected infection was defined as at least two systemic antimicrobial records within 24 h (the first record being the time of suspected infection), with the SOFA and onset rules unchanged. Hourly features were rebuilt with the frozen feature code and the concept and unit mapping from MIMIC-IV to eICU-CRD, and the 31-concept event stream was built with the same mapping; the 99 fluid and blood-product columns without a source in eICU-CRD were treated as missing with missingness indicators; preprocessing statistics, model weights, calibrators and alarm thresholds were all carried over from MIMIC-IV, and nothing was fitted on eICU-CRD. Local recalibration was a logistic recalibration fitted on eICU-CRD with 5-fold cross-fitting by patient, used only to show that calibration can be restored locally, and the 16 published algorithms were evaluated on eICU-CRD by the same rules (Supplementary Fig. 5d; Supplementary Table 23).

**Published subtypes used as comparators**

Onset comparator subtypes were re-derived on the discovery set with the original methods and frozen for test evaluation (Supplementary Table 25): the α–δ phenotypes of Seymour et al.[28] (consensus k-means, K = 4), the coagulation–inflammation subphenotypes of SepsisFormer[29] (Gaussian mixture, K = 2 on complete cases), the latent profile subtypes of Zhang et al.[50] (K = 4), the consensus clusters of Boussina et al.[30], and hidden-state clusters of a feed-forward network trained on the hourly features (K = 3). Course comparator subtypes were likewise derived on the discovery set (Supplementary Tables 4 and 31): the 72-h SOFA-trajectory subtypes of Xu et al.[33], the temperature-trajectory subtypes of Bhavani et al.[32] and latent-class organ-support trajectories.

**Prognostic analyses**

We evaluated onset contexts from $T_0$ and early courses from $T_0$ + 72 h[35] (891 test patients still in hospital; endpoint: 28-day in-hospital death with live discharge competing[36]; Supplementary Table 20). Aalen–Johansen curves were compared by Gray's test, and cause-specific Cox models gave hazard ratios before adjustment, after adjustment for age, sex, onset SOFA, vasopressors, invasive ventilation and onset context, and after adding worst SOFA at 48–72 h. A Fine–Gray subdistribution model, exclusion of the cardiac vascular ICU and 28-day all-cause death were sensitivity analyses (Supplementary Tables 27–29). The prognostic increment of a classification was measured as the difference (ΔC) in Harrell's C-index and in the Wolbers C-index for competing risks[45], with coefficients fitted on the discovery set and applied to the test set (Supplementary Table 22).

**Statistical analysis**

In all tables and figures, we did not treat repeated hourly scores from the same patient as independent. Instead, we drew 2,000 patient-cluster bootstrap resamples with a single random seed (42) shared across all models so that every model-to-model difference came from the exact same resampled patients (and resampled hospitals in eICU-CRD). From those resamples we obtained 95% percentile intervals for AUROC, AUPRC, the Platt-calibrated slope, intercept and observed-to-expected ratio, and decision-curve net benefit between 2% and 30% threshold probability (reporting 10% in Table 2), whereas single binomial proportions in Fig. 5c, Fig. 6b and Supplementary Table 3 used Wilson intervals. Agreement between classifications was measured by the adjusted Rand index, Cohen's κ and bias-corrected Cramér's V. Multiple comparisons were adjusted within pre-defined families by the Holm method; under the analysis registration, statements requiring every comparison to hold, such as "for each of the 16 algorithms", are intersection–union tests and were not adjusted further. All calculations used Python 3.10 with PyTorch, scikit-learn, LightGBM, SHAP and lifelines, together with R 4.5.3.

**Use of the internal test set and analysis registration**

Each model configuration was evaluated once on the internal test set using parameters fixed on the development, validation or discovery sets. The benchmark comparisons, ablations, encoder probes and trajectory analyses were run according to a timestamped analysis registration. Two validation-gated updates are documented in Supplementary Table 18: removing the 19 respiratory-support features with future timestamps and testing patient-state components in the risk branch (additional post hoc comparisons in Supplementary Tables 2, 13, 21 and 22).

**Scope of the study**

The risk outputs of SepsisFlow are limited to the probabilities of onset within 6, 12 and 24 h at each scoring hour and the first alarm derived from them; it outputs no uncertainty interval or multi-day organ forecast, and the three early courses classify the records observed in the 72 h after onset. In feasibility analyses on MIMIC-IV, estimating treatment-effect heterogeneity by onset context or early course (for intravenous hydrocortisone, red-cell transfusion, vasopressin, antimicrobial de-escalation and loop diuretics) was underpowered or confounded by indication, so this study does not answer questions of treatment choice.

**Data availability**

All data analysed in this study are publicly available from their original publishers. MIMIC-IV (version 3.1; https://physionet.org/content/mimiciv/) and the eICU Collaborative Research Database (version 2.0; https://physionet.org/content/eicu-crd/) are hosted on PhysioNet under credentialed data use agreements. The pretrained event-stream encoder weights were released by Wornow et al. (https://huggingface.co/StanfordShahLab/mamba-tiny-4096-clmbr).

**Code availability**

The frozen model weights and inference code of SepsisFlow are available on GitHub (https://github.com/WanzheUTokyo/SepsisFlow), with installation instructions, usage examples and documentation.

**Acknowledgements**

Computational resources for model inference and device testing were provided by the SHIROKANE supercomputer system at the Human Genome Center, The Institute of Medical Science, The University of Tokyo. Model training and data analyses were performed using the Supermicro ARS-111GL-DNHR-LCC and FUJITSU Server PRIMERGY CX2550 M7 (Miyabi) systems at the Joint Center for Advanced High-Performance Computing (JCAHPC).

This work was supported by the Japan Science and Technology Agency (JST) SPRING program (grant no. JPMJSP2108).

**Author contributions**

Y.Z. and W.X. contributed equally to this work.

Conceptualization: Y.Z and W.X.; methodology: Y.Y. and W.X.; formal analysis: Y.Y., W.X., and Y.Y.; investigation: Y.Z., and Y.Y.; data curation: Y.Y.; statistical review: Y.Y., Z.Z., and W.X.; original manuscript writing: Y.Z; manuscript review and editing: W.X., Z.Z., Y.Z, J.W., Y.Y., and K.N.; supervision: K.N. All authors read and approved the final manuscript.

**Competing interests**

The authors declare no competing interests.

## Abbreviations

**ARI**: adjusted Rand index
**AUPRC**: area under the precision–recall curve
**AUROC**: area under the receiver operating characteristic curve
**CI**: confidence interval
**CNS**: central nervous system
**CSICU / CTICU**: cardiac surgical / cardiothoracic intensive care unit
**CVICU**: cardiac vascular intensive care unit
**DCA**: decision curve analysis
**EASP**: explainable artificial intelligence sepsis predictor
**EHR**: electronic health record
**eICU-CRD**: eICU Collaborative Research Database
**$FiO_2$**: fraction of inspired oxygen
**GCS**: Glasgow Coma Scale
**GRU**: gated recurrent unit
**HR**: hazard ratio
**ICU**: intensive care unit
**INR**: international normalized ratio
**IQR**: interquartile range

**KL**: Kullback–Leibler divergence

**LSTM**: long short-term memory

**MAP**: mean arterial pressure

**MIMIC-IV**: Medical Information Mart for Intensive Care IV

**NCL**: neighborhood contrastive learning

**NEWS2**: National Early Warning Score 2

**O/E**: observed-to-expected ratio

**$PaCO_2$**: arterial partial pressure of carbon dioxide

**$PaO_2$**: arterial partial pressure of oxygen

**PCA**: principal component analysis

**PEEP**: positive end-expiratory pressure

**PPV**: positive predictive value

**PTT**: partial thromboplastin time

**qSOFA**: quick Sequential Organ Failure Assessment

**ROC**: receiver operating characteristic

**RRT**: renal replacement therapy

**SeFT**: set functions for time series

**SHAP**: SHapley Additive exPlanations

**SIRS**: systemic inflammatory response syndrome

**SOFA**: Sequential Organ Failure Assessment

**$SpO_2$**: peripheral oxygen saturation

**$T_0$**: time of sepsis onset

**TCN**: temporal convolutional network

**TreeSHAP**: tree-based SHapley Additive exPlanations

**YAIB**: Yet Another ICU Benchmark

## Figure legends

**Fig. 1 | SepsisFlow: one hourly clinical clock, two branches. a,** Data. MIMIC-IV was used for development and internal evaluation; the frozen prediction model was applied without retraining to 22 eICU-CRD hospitals (Supplementary Fig. 5). **b,** Study timeline. Before onset the question is who will develop sepsis: hourly 6-, 12- and 24-h risk, with a diamond marking the first alarm. At onset ($T_0$) the onset context is assigned; from $T_0$ to $T_0$ + 72 h the early course is observed, and outcomes are followed to day 28. Scoring starts at ICU admission, and stays with onset <4 h after admission are excluded. **c,** SepsisFlow. Top row: hourly risk prediction, in which a frozen pretrained Mamba event-stream encoder, summarised into 32 principal components, and 1,678 hourly features (1,710 inputs in all) are combined in gradient-boosted trees. Bottom row: the dynamic patient state h(t) and its three outputs—onset context, state trajectories and 24-h organ-level risk. Right column: the observed early course, derived from the trajectories recorded up to 72 h after onset with a separate sequence encoder. The event encoder processes measurement tokens of 31 clinical concepts and no treatment, antibiotic, culture or time-gap tokens. Only inputs available at each hour are used, and antibiotics and cultures are excluded.

**Fig. 2 | When SepsisFlow warns.** MIMIC-IV test set: 4,161 stays, 997 cases, 182,415 hourly scores (630 cases [63%] at 4–11 h). **a,** ROC curves at 24 h (vermillion solid; AUROC 0.822, 0.807–0.836), 12 h (orange dashed; 0.843) and 6 h (light orange dotted; 0.859). **b,** Precision–recall curves for the same horizons (AUPRC 0.292, 0.289 and 0.284; grey lines, prevalence 0.067, 0.048 and 0.032). **c,** Cumulative share of the 997 cases with a first alarm (vermillion solid; threshold 0.187 uncalibrated, 0.191 calibrated) and a first antibiotic order (reddish purple dashed) by hour from onset (ribbons, 95% CIs). 847 cases (85.0%) were alarmed

before onset (median −5.7 h, −6.2 to −5.4). Of 702 cases alarmed 0–24 h before onset, 555 (79.1%, 76.1–82.1) preceded the first order by a median of 3.0 h (2.6–3.4); 416 (59.3%) came at the first ICU hour. **d,** Share of the 997 cases first alarmed 0–24 h before onset against alarm burden (first alarms per 100 patient-days at risk; 7,513 patient-days), with at most one alarm per stay. Vermillion solid, hourly re-scoring; light orange dashed, first assessment only. At the validation load of 0.31 first alarms per patient-day, hourly re-scoring detects 70.4% and first assessment only 62.3% (+8.1 percentage points, 5.7–10.5). Markers with 95% CI crosses: NEWS2 ≥ 5 (blue filled square), NEWS2 ≥ 7 (blue open square), qSOFA ≥ 2 (green triangle) and SIRS ≥ 2 (orange diamond); the grey dashed line marks the NEWS2 ≥ 5 burden (46.0). At that burden NEWS2 ≥ 5 detects 56.8% (53.7–59.8), the context-only reference 72.9% (70.0–75.6), and SepsisFlow 76.4% (73.7–79.1; +19.7 points vs NEWS2 ≥ 5 and +3.5 vs reference; Supplementary Table 32). 95% CIs from 2,000 patient-cluster bootstrap replicates.

**Fig. 3 | SepsisFlow against 16 published algorithms on the same data.** MIMIC-IV internal test set (4,161 stays, 182,415 hourly predictions, 997 cases; sepsis onset within 24 h). The 16 published algorithms (10 studies) were retrained with official code on the same cohort, labels and hourly protocol. In **a–e**, SepsisFlow is vermillion (solid), the three strongest comparators are Morrill signature + LightGBM (sky blue, dashed), EASP XGBoost (reddish purple, dotted) and SeFT-Attn (blue, dot-dash), and the other 13 are light grey. **a,** ROC curves (AUROC: SepsisFlow 0.822, Morrill 0.808, EASP 0.801, SeFT-Attn 0.792). **b,** Precision–recall curves; dashed line, prevalence (6.7%; AUPRC: SepsisFlow 0.292, Morrill 0.273). **c,** Calibration after Platt scaling fitted on each model's validation predictions (restricted cubic spline curves between the 0.5th and 99th percentiles); vermillion band, SepsisFlow 95% CI; black dashed, ideal calibration. Calibration slope: SepsisFlow 1.01 (0.95–1.08); published algorithms 0.90–1.17. **d,** Decision curves of the recalibrated predictions (threshold probability 2%–30%) against treat all (grey dashed) and treat none (black); net benefit at 10%: SepsisFlow 0.0240 (0.0206–0.0275). **e,** Alarm-burden curves of all 17 models; vertical dashed line marks the NEWS2 ≥ 5 burden (46.0 first alarms per 100 patient-days). **f,** Paired differences, SepsisFlow minus each algorithm, in 24-h AUROC (left) and AUPRC (right) with 95% CIs from 2,000 patient-cluster bootstrap replicates. ΔAUROC ranges from +0.014 to +0.075 and ΔAUPRC from +0.019 to +0.125; all 32 intervals exclude 0 (Holm-adjusted P = 0.016 for each).

**Fig. 4 | Physiological signals in SepsisFlow's warnings rise towards onset, relative to patients at the same hour of their ICU stay.** Values are exact interventional TreeSHAP attributions of the SepsisFlow 24-h output (log-odds) on the MIMIC-IV validation set (8,464 stays, 1,960 cases); each hourly score is explained against 50 development-set scores at the

same hour since ICU admission, giving ICU time zero attribution. Seven organ concepts (respiratory, metabolism & inflammation, neurology, circulation, kidney, blood & coagulation, liver) sum to Physiology; the other concepts are observation process, static context, fluid therapy, fluid removal, vasopressors and event-stream components (Supplementary Table 33). Bars, 95% CIs from 2,000 patient bootstrap replicates. **a,** SHAP values of the 15 vital-sign, GCS and laboratory inputs with the largest mean |SHAP| for 3,960 hourly scores (1,960 cases at their last score before onset and one random score from 2,000 stays without sepsis; blue low, vermillion high, grey unmeasured; Supplementary Fig. 3a). The contribution of lowest MAP is highest just above 70 mmHg, the cardiovascular SOFA cut-off. **b,** Within-patient change in the SHAP value of each concept from >24 h to the last 6 h before onset in the 427 cases scored in both windows (organ concepts vermillion, observation process reddish purple, others black). **c,** Mean SHAP value of Physiology (vermillion, solid) and the observation process (reddish purple, dotted) by time-to-onset bin in the 427 cases. Grey dashed line with points, Physiology in five matched patients without sepsis per case (1,277 stays; case–control change contrast +0.635, 0.562–0.710; Supplementary Table 34). **d,** Share of each organ concept in the positive physiological SHAP at the last score before onset by onset context in 1,581 validation cases (Late ICU onset 568, Early onset, non-CVICU 676, Early onset, CVICU 331).

**Fig. 5 | Patient state at onset and its three onset contexts.** MIMIC-IV internal test set (n = 997: Late ICU onset 359, Early onset, non-CVICU 407, Early onset, CVICU 231). **a,** FastICA two-dimensional projection (fitted on the discovery set) of the last pre-onset patient state $h(T_0-)$, with kernel-density contours at 30% and 65% of each context peak. **b,** Two-question bedside rule (Yes branches right): onset <12 h after ICU admission? (No: Late ICU onset); first ICU unit is CVICU? (No: Early onset, non-CVICU; Yes: Early onset, CVICU). Leaf nodes show the 995 classified patients and crude 28-day in-hospital deaths among 990 with known status (59/364 [16.2%], 65/413 [15.7%], 2/213 [0.9%]); rule agreement with model contexts is 87.9% (875/995; Cohen's $\kappa$ = 0.81). **c,** Proportion of patients with SOFA component ≥2 at onset by organ system and context (bars show Wilson 95% CIs). **d,** Aalen–Johansen cumulative incidence of in-hospital death to day 28 with live discharge as a competing event (shaded 95% CIs). Day-28 mortality was 17.1% (13.4–21.2) in Late ICU onset, 15.1% (11.8–18.7) in Early onset, non-CVICU and 2.2% (0.8–4.8) in Early onset, CVICU (Gray's test $P = 2.4 \times 10^{-7}$; n = 356, 405 and 231). **e,** Alluvial flow of 995 patients from the two-question rule to onset contexts and Seymour α–δ phenotypes (Cramér's V = 0.50,

ARI = 0.18). Early onset, CVICU is 87% α (201/231) and Late ICU onset is 54% γ (194/359), reflecting perioperative cardiac surgery stays meeting Sepsis-3 criteria.

**Fig. 6 | Early courses after onset.** Test-set patients alive and in hospital at the 72-h landmark, n = 891 (De-escalating 339, Early ICU discharge 255, Persistent organ support 297); 84 in-hospital deaths by day 28. Vermillion denotes Persistent organ support in **a–d** and **f** and SepsisFlow in **e** and **g**. **a,** Flows from onset context to early course to status at day 28; flows into 'Died': Persistent organ support 64/297, De-escalating 15/339, Early ICU discharge 5/255. **b,** In-hospital death by day 28 by context and course (bars, Wilson 95% intervals; numbers below the bars, patients): Persistent organ support 21.5% (17.3–26.6), De-escalating 4.4% (2.7–7.2), Early ICU discharge 2.0% (0.8–4.5); within Late ICU onset 24.3% vs 4.7% and within Early onset, non-CVICU 20.4% vs 4.3%; bars with fewer than 20 patients are drawn lighter with grey error bars. **c,** Hazard ratio of in-hospital death by day 28 for Persistent organ support versus De-escalating from the 72-h landmark (cause-specific Cox models with live discharge competing; Wald 95% CIs; log scale). Open circle: unadjusted, 3.12 (1.77–5.49). Filled circles: adjusted for age, sex, SOFA score, vasopressor and invasive ventilation at onset and onset context, 4.10 (2.15–7.81); further adjusted for SOFA at 72 h, 2.80 (1.42–5.55). **d,** Hourly means after onset by course: noradrenaline-equivalent dose, share on advanced respiratory support, share on renal replacement therapy and mean GCS; lines drawn where ≥ 20 patients are observed, ribbons pointwise 95% intervals (Supplementary Fig. 4e). **e,** Test-set AUROC for a new SOFA component ≥ 2 within 24 h from hourly points 0–72 h after onset (889–955 patients per organ). Vermillion filled points, linear model on the patient state; grey open points, the same model on the six current SOFA components. Δ (Holm-adjusted 95% CI): respiratory +0.193 (+0.138 to +0.248), coagulation −0.005 (−0.038 to +0.027), liver +0.048 (−0.050 to +0.147), cardiovascular +0.143 (+0.056 to +0.232), CNS +0.131 (+0.076 to +0.189), renal +0.130 (+0.056 to +0.201). **f,** Patient state of 995 test patients (De-escalating 404, Early ICU discharge 260, Persistent organ support 331) in the Fig. 5a projection and axes; lines, mean positions per 6-h bin from 6–0 h before onset to 72 h after it; grey streamlines, grid-averaged hour-to-hour displacement (65,892 pairs). **g,** AUROC of a discovery-fitted linear model on the state for Persistent organ support vs the other two courses, per 6-h bin from 24 h before to 24 h after onset. Vermillion solid line, all patients in the bin: 0.795 (0.737–0.846) at 24–18 h before onset, rising to 0.896 (0.874–0.916) at 6–0 h (vs 0.860 for the bedside model; Supplementary Table 21); light orange dashed line, the fixed cohort of 278 patients observed ≥ 24 h before onset (0.795 to 0.863).

# Figures

## Fig. 1

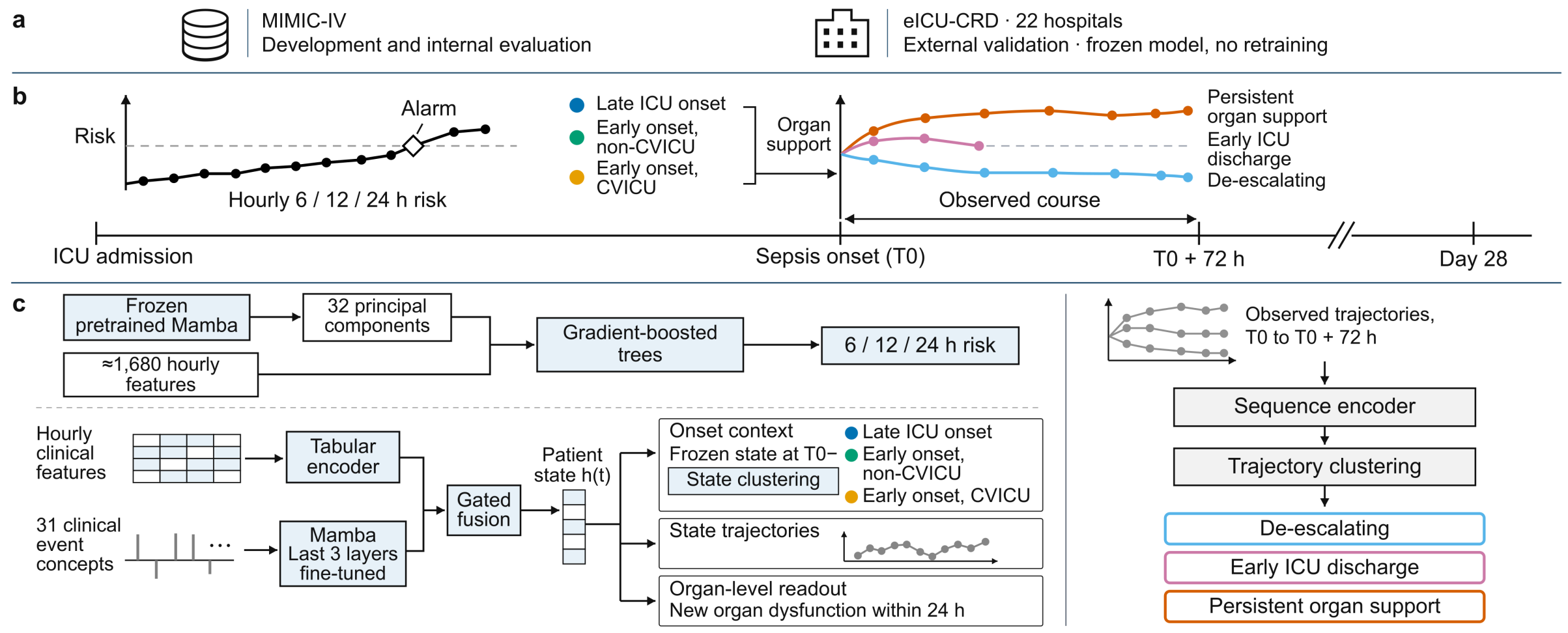

## Fig. 2

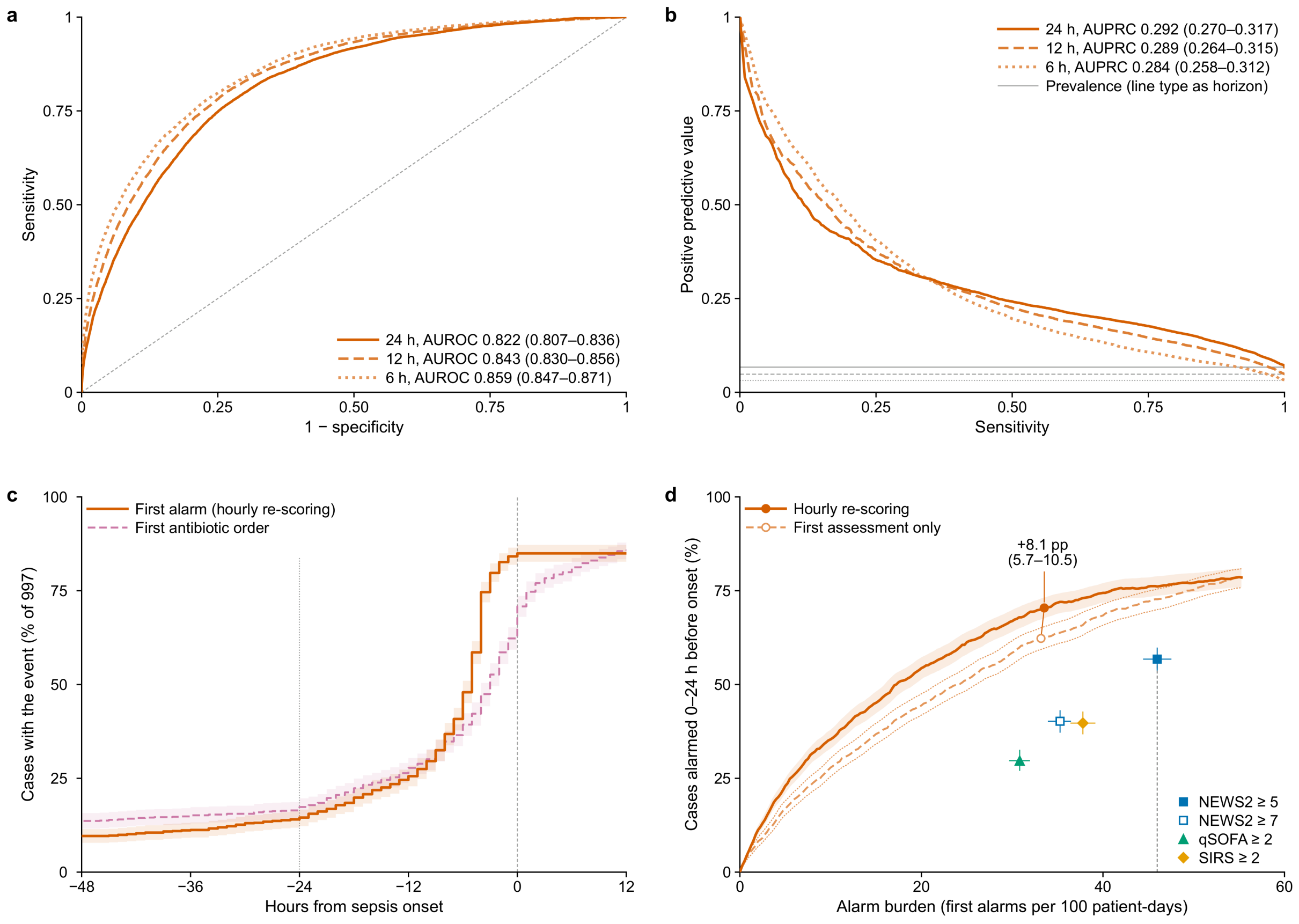

## Fig. 3

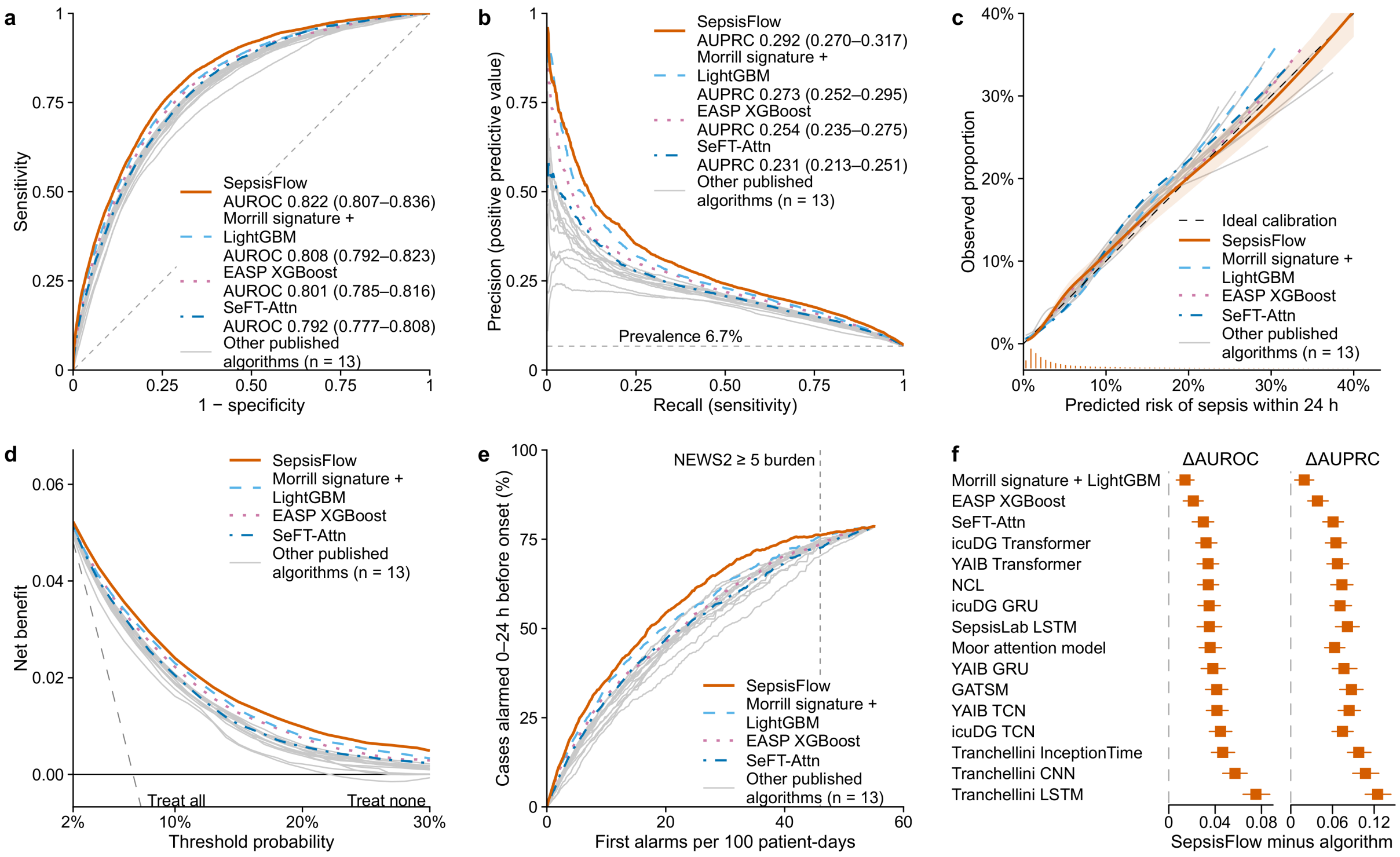
a
Sensitivity
1 − specificity
SepsisFlow AUROC 0.822 (0.807–0.836)
Morrill signature + LightGBM AUROC 0.808 (0.792–0.823)
EASP XGBoost AUROC 0.801 (0.785–0.816)
SeFT-Attn AUROC 0.792 (0.777–0.808)
Other published algorithms (n = 13)
b
Precision (positive predictive value)
Recall (sensitivity)
SepsisFlow AUPRC 0.292 (0.270–0.317)
Morrill signature + LightGBM AUPRC 0.273 (0.252–0.295)
EASP XGBoost AUPRC 0.254 (0.235–0.275)
SeFT-Attn AUPRC 0.231 (0.213–0.251)
Other published algorithms (n = 13)
Prevalence 6.7%
c
Observed proportion
Predicted risk of sepsis within 24 h
Ideal calibration
SepsisFlow
Morrill signature + LightGBM
EASP XGBoost
SeFT-Attn
Other published algorithms (n = 13)
d
Net benefit
Threshold probability
Treat all
Treat none
e
Cases alarmed 0–24 h before onset (%)
First alarms per 100 patient-days
NEWS2 ≥ 5 burden
f
ΔAUROC
ΔAUPRC
Morrill signature + LightGBM
EASP XGBoost
SeFT-Attn
icuDG Transformer
YAIB Transformer
NCL
icuDG GRU
SepsisLab LSTM
Moor attention model
YAIB GRU
GATSM
YAIB TCN
icuDG TCN
Tranchellini InceptionTime
Tranchellini CNN
Tranchellini LSTM
SepsisFlow minus algorithm

**Fig. 4**

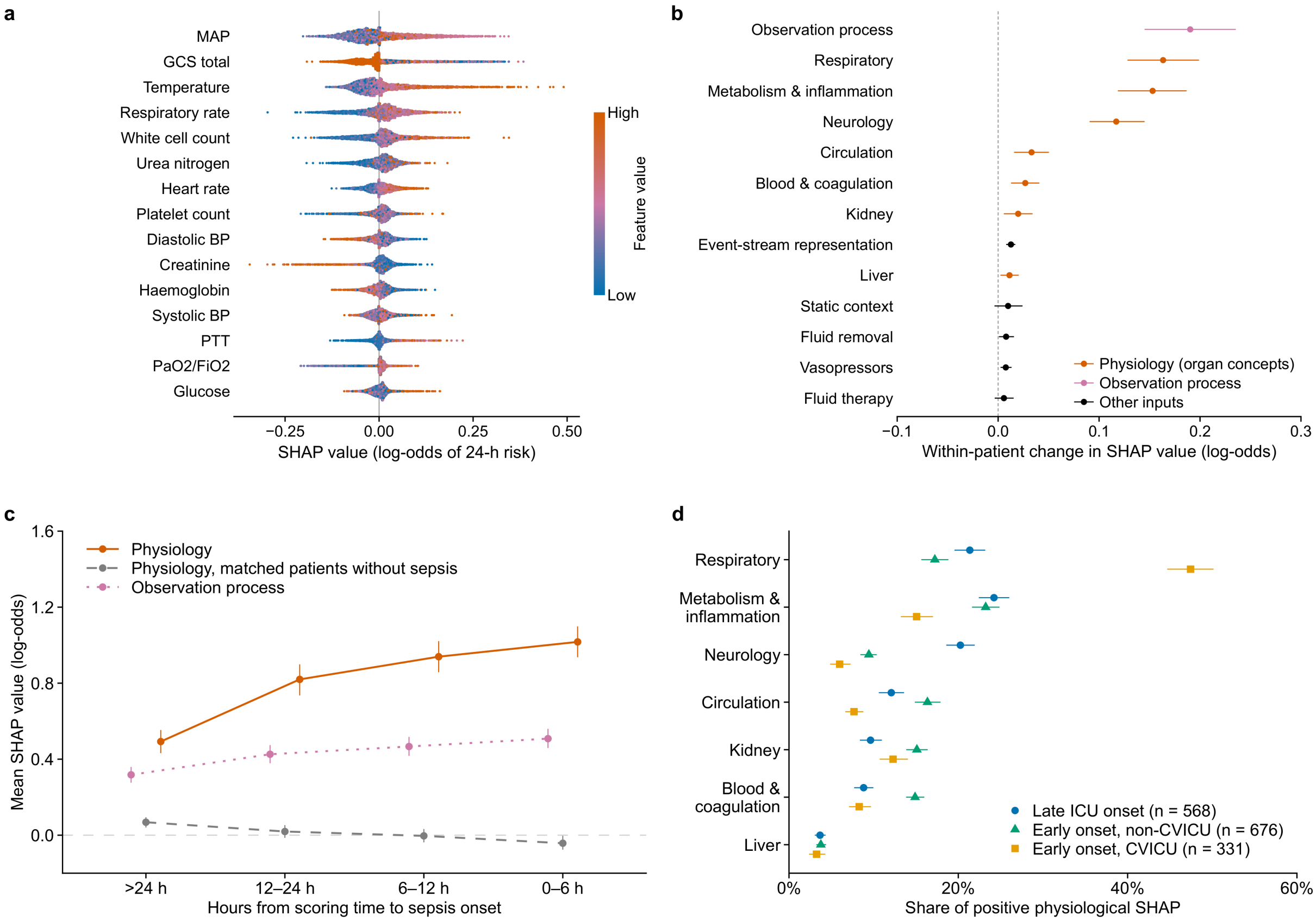
a
MAP
GCS total
Temperature
Respiratory rate
White cell count
Urea nitrogen
Heart rate
Platelet count
Diastolic BP
Creatinine
Haemoglobin
Systolic BP
PTT
PaO2/FiO2
Glucose
SHAP value (log-odds of 24-h risk)
High
Feature value
Low
b
Observation process
Respiratory
Metabolism & inflammation
Neurology
Circulation
Blood & coagulation
Kidney
Event-stream representation
Liver
Static context
Fluid removal
Vasopressors
Fluid therapy
Physiology (organ concepts)
Observation process
Other inputs
Within-patient change in SHAP value (log-odds)
c
Mean SHAP value (log-odds)
Physiology
Physiology, matched patients without sepsis
Observation process
>24 h
12–24 h
6–12 h
0–6 h
Hours from scoring time to sepsis onset
d
Respiratory
Metabolism & inflammation
Neurology
Circulation
Kidney
Blood & coagulation
Liver
Late ICU onset (n = 568)
Early onset, non-CVICU (n = 676)
Early onset, CVICU (n = 331)
Share of positive physiological SHAP

**Fig. 5**

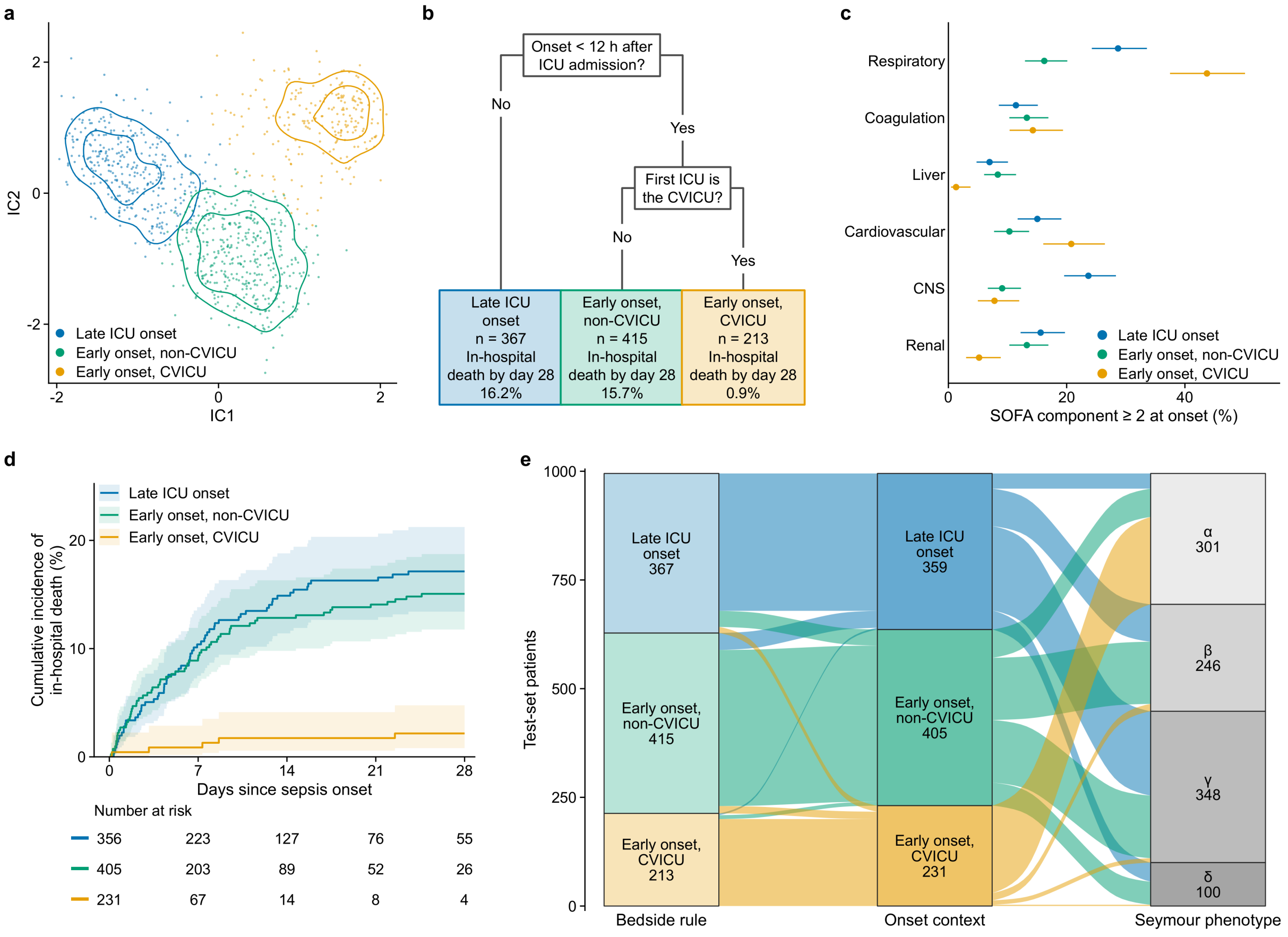
a
IC2
IC1
Late ICU onset
Early onset, non-CVICU
Early onset, CVICU
b
Onset < 12 h after ICU admission?
No
Yes
First ICU is the CVICU?
No
Yes
Late ICU onset n = 367 In-hospital death by day 28 16.2%
Early onset, non-CVICU n = 415 In-hospital death by day 28 15.7%
Early onset, CVICU n = 213 In-hospital death by day 28 0.9%
c
Respiratory
Coagulation
Liver
Cardiovascular
CNS
Renal
Late ICU onset
Early onset, non-CVICU
Early onset, CVICU
SOFA component ≥ 2 at onset (%)
d
Cumulative incidence of in-hospital death (%)
Late ICU onset
Early onset, non-CVICU
Early onset, CVICU
Days since sepsis onset
Number at risk
356 223 127 76 55
405 203 89 52 26
231 67 14 8 4
e
Test-set patients
Late ICU onset 367
Early onset, non-CVICU 415
Early onset, CVICU 213
Late ICU onset 359
Early onset, non-CVICU 405
Early onset, CVICU 231
α 301
β 246
γ 348
δ 100
Bedside rule
Onset context
Seymour phenotype

**Fig. 6**

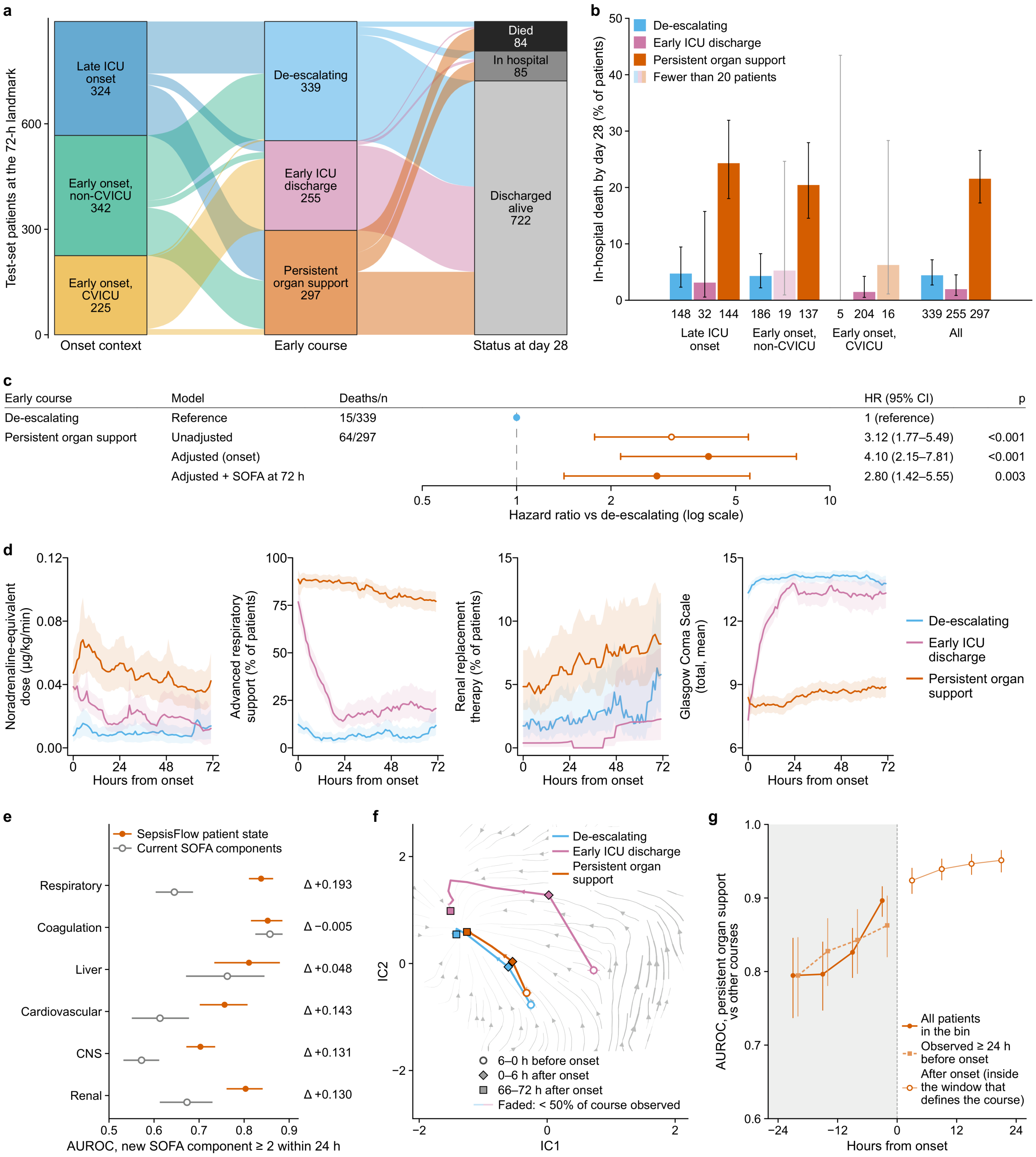
a
Test-set patients at the 72-h landmark
600
300
0
Late ICU onset 324
Early onset, non-CVICU 342
Early onset, CVICU 225
De-escalating 339
Early ICU discharge 255
Persistent organ support 297
Died 84
In hospital 85
Discharged alive 722
Onset context
Early course
Status at day 28
b
In-hospital death by day 28 (% of patients)
De-escalating
Early ICU discharge
Persistent organ support
Fewer than 20 patients
148 32 144
Late ICU onset
186 19 137
Early onset, non-CVICU
5 204 16
Early onset, CVICU
339 255 297
All
c
Early course
Model
Deaths/n
HR (95% CI)
p
De-escalating
Reference
15/339
1 (reference)
Persistent organ support
Unadjusted
64/297
3.12 (1.77–5.49)
<0.001
Adjusted (onset)
4.10 (2.15–7.81)
<0.001
Adjusted + SOFA at 72 h
2.80 (1.42–5.55)
0.003
Hazard ratio vs de-escalating (log scale)
d
Noradrenaline-equivalent dose (µg/kg/min)
Advanced respiratory support (% of patients)
Renal replacement therapy (% of patients)
Glasgow Coma Scale (total, mean)
Hours from onset
De-escalating
Early ICU discharge
Persistent organ support
e
SepsisFlow patient state
Current SOFA components
Respiratory
Δ +0.193
Coagulation
Δ −0.005
Liver
Δ +0.048
Cardiovascular
Δ +0.143
CNS
Δ +0.131
Renal
Δ +0.130
AUROC, new SOFA component ≥ 2 within 24 h
f
De-escalating
Early ICU discharge
Persistent organ support
6–0 h before onset
0–6 h after onset
66–72 h after onset
Faded: < 50% of course observed
IC1
IC2
g
AUROC, persistent organ support vs other courses
All patients in the bin
Observed ≥ 24 h before onset
After onset (inside the window that defines the course)
Hours from onset

# Tables

## Table 1 | Baseline characteristics of the development, validation, internal test and external validation cohorts

| Characteristic | MIMIC-IV development set | MIMIC-IV validation set | MIMIC-IV internal test set | Internal test set: sepsis | Internal test set: no sepsis | eICU-CRD, 22 hospitals |
|---|---|---|---|---|---|---|
| ICU stays, n | 29,403 | 8,464 | 4,161 | 997 | 3,164 | 12,433 |
| Patients, n | 29,403 | 8,464 | 4,161 | 997 | 3,164 | 12,433 |
| Sepsis within the ICU stay (onset ≥ 4 h after admission), n (%) | 6,791 (23.1) | 1,960 (23.2) | 997 (24.0) | 997 (100.0) | 0 (0.0) | 421 (3.4) |
| Hourly scores, total n | 1,339,337 | 379,273 | 182,415 | 24,784 | 157,631 | 738,424 |
| Hourly scores per stay, median (IQR) | 30 (17–56) | 30 (17–55) | 30 (17–53) | 9 (6–20) | 36 (23–60) | 39 (21–70) |
| Age, years, median (IQR) | 65 (53–77) | 65 (53–76) | 66 (54–77) | 67 (55–77) | 66 (53–77) | 65 (53–77) |
| Female sex, n (%) | 12,700 (43.2) | 3,741 (44.2) | 1,892 (45.5) | 431 (43.2) | 1,461 (46.2) | 5,728 (46.1) |
| Race and ethnicity, n (%) | | | | | | |
| White | 19,248 (65.5) | 5,591 (66.1) | 2,735 (65.7) | 632 (63.4) | 2,103 (66.5) | 10,172 (81.8) |
| Black | 2,427 (8.3) | 705 (8.3) | 317 (7.6) | 71 (7.1) | 246 (7.8) | 1,608 (12.9) |
| Hispanic or Latino | 1,090 (3.7) | 287 (3.4) | 139 (3.3) | 37 (3.7) | 102 (3.2) | 97 (0.8) |
| Asian | 873 (3.0) | 243 (2.9) | 124 (3.0) | 27 (2.7) | 97 (3.1) | 111 (0.9) |
| Other | 1,126 (3.8) | 298 (3.5) | 175 (4.2) | 46 (4.6) | 129 (4.1) | 432 (3.5) |
| Unknown or not reported | 4,639 (15.8) | 1,340 (15.8) | 671 (16.1) | 184 (18.5) | 487 (15.4) | 13 (0.1) |
| Hospital admission type, n (%) | | | | | | |
| Emergency or urgent | 20,001 (68.0) | 5,793 (68.4) | 2,828 (68.0) | 721 (72.3) | 2,107 (66.6) | not recorded |
| Elective or same-day surgical | 4,862 (16.5) | 1,410 (16.7) | 707 (17.0) | 162 (16.2) | 545 (17.2) | not recorded |
| Observation | 4,540 (15.4) | 1,261 (14.9) | 626 (15.0) | 114 (11.4) | 512 (16.2) | not recorded |
| First ICU unit, n (%) | | | | | | |
| Medical ICU | 4,415 (15.0) | 1,269 (15.0) | 620 (14.9) | 159 (15.9) | 461 (14.6) | 1,004 (8.1) |
| Surgical ICU | 4,335 (14.7) | 1,297 (15.3) | 608 (14.6) | 134 (13.4) | 474 (15.0) | 893 (7.2) |
| Medical/surgical ICU | 3,850 (13.1) | 1,054 (12.5) | 529 (12.7) | 134 (13.4) | 395 (12.5) | 7,861 (63.2) |
| Trauma surgical ICU | 3,925 (13.3) | 1,118 (13.2) | 573 (13.8) | 121 (12.1) | 452 (14.3) | 0 (0.0) |
| Coronary care unit | 3,494 (11.9) | 1,021 (12.1) | 503 (12.1) | 88 (8.8) | 415 (13.1) | 752 (6.0) |
| Cardiac vascular ICU (CVICU) | 5,496 (18.7) | 1,591 (18.8) | 799 (19.2) | 264 (26.5) | 535 (16.9) | 0 (0.0) |
| Mixed coronary care/cardiothoracic ICU | 0 (0.0) | 0 (0.0) | 0 (0.0) | 0 (0.0) | 0 (0.0) | 1,137 (9.1) |
| Neurological ICU or unit | 3,831 (13.0) | 1,084 (12.8) | 518 (12.4) | 90 (9.0) | 428 (13.5) | 786 (6.3) |
| Other | 57 (0.2) | 30 (0.4) | 11 (0.3) | <15 | <15 | 0 (0.0) |
| Charlson comorbidity index (age points excluded), median (IQR) | 2 (1–3) | 2 (1–3) | 2 (1–4) | 2 (1–4) | 2 (1–3) | not derived |
| not available, n (%) | 0 | 0 | 0 | 0 | 0 | — |
| SOFA score at ICU admission (24 h up to the first ICU hour), median (IQR) | 0 (0–2) | 0 (0–2) | 1 (0–2) | 1 (0–2) | 0 (0–2) | not derived |
| not available, n (%) | 10 (0.0) | <5 | <5 | 0 | <5 | — |
| Hours from ICU admission to sepsis onset (cases), median (IQR) | 8.4 (5.5–20.3) | 8.7 (5.5–20.2) | 8.2 (5.3–19.9) | 8.2 (5.3–19.9) | — | 22.7 (7.0–61.7) |
| ICU length of stay, days, median (IQR) | 1.8 (1.0–3.2) | 1.7 (1.0–3.2) | 1.7 (1.0–3.1) | 3.5 (1.5–7.3) | 1.5 (1.0–2.5) | 1.7 (0.9–3.0) |
| Hospital length of stay, days, median (IQR) | 6.1 (3.6–10.6) | 6.1 (3.6–10.5) | 6.3 (3.6–10.6) | 8.9 (5.3–15.9) | 5.7 (3.1–9.1) | 4.9 (2.6–8.8) |
| In-hospital death, n (%) | 2,263 (7.7) | 673 (8.0) | 375 (9.0) | 139 (13.9) | 236 (7.5) | 1,094 (9.1) |
| not available, n (%) | 0 | 0 | 0 | 0 | 0 | 389 (3.1) |

Values are n (%) or median (IQR); counts of 1–4 are shown as <5 (<15 where required to prevent complementary disclosure). Cohort comprises adult first ICU stays scored hourly from admission, excluding sepsis onset <4 h after admission. Baseline SOFA reflects the first ICU hour; in eICU-CRD, suspected infection required ≥2 systemic antimicrobial records within 24 h.

## Table 2 | Comparison of SepsisFlow with 16 published algorithms on the MIMIC-IV internal test set (24-h horizon)

| Model | Study | AUROC | ΔAUROC | AUPRC | ΔAUPRC | Calibration slope | Calibration-in-the-large | Net benefit at 10% | Detection at NEWS2 ≥ 5 burden, % | Median lead time, h | AUROC before in-ICU treatment | AUPRC before in-ICU treatment | ΔAUPRC before in-ICU treatment |
|---|---|---|---|---|---|---|---|---|---|---|---|---|---|
| SepsisFlow | this study | 0.822 (0.807 to 0.836) | — | 0.292 (0.270 to 0.317) | — | 1.01 (0.95 to 1.08) | 0.040 (-0.053 to 0.126) | 0.0240 (0.0206 to 0.0275) | 76.4 (73.7 to 79.1) | 6.4 (6.1 to 6.7) | 0.834 (0.811 to 0.855) | 0.335 (0.306 to 0.367) | — |
| Morrill signature + LightGBM | Morrill et al., Crit Care Med 2020 (PhysioNet/CinC 2019 winner) | 0.808 (0.792 to 0.823) | +0.014 (+0.006 to +0.022) | 0.273 (0.252 to 0.295) | +0.019 (+0.006 to +0.033) | 1.07 (0.96 to 1.17) | 0.040 (-0.050 to 0.123) | 0.0230 (0.0194 to 0.0266) | 75.9 (73.2 to 78.6) | 6.4 (5.9 to 6.7) | 0.823 (0.800 to 0.847) | 0.310 (0.282 to 0.341) | +0.026 (+0.006 to +0.045) |
| EASP XGBoost | Yang et al., Crit Care Med 2020 (EASP) | 0.801 (0.785 to 0.816) | +0.021 (+0.012 to +0.030) | 0.254 (0.235 to 0.275) | +0.038 (+0.024 to +0.054) | 1.04 (0.97 to 1.11) | 0.039 (-0.049 to 0.119) | 0.0222 (0.0190 to 0.0255) | 73.7 (70.9 to 76.5) | 6.4 (6.0 to 6.7) | 0.811 (0.785 to 0.837) | 0.286 (0.259 to 0.316) | +0.049 (+0.029 to +0.070) |
| SeFT-Attn | Horn et al., ICML 2020 (SeFT) | 0.792 (0.777 to 0.808) | +0.030 (+0.020 to +0.039) | 0.231 (0.213 to 0.251) | +0.061 (+0.046 to +0.076) | 1.11 (1.03 to 1.19) | 0.053 (-0.034 to 0.129) | 0.0205 (0.0173 to 0.0237) | 73.4 (70.6 to 76.1) | 6.4 (6.1 to 6.7) | 0.804 (0.780 to 0.828) | 0.250 (0.228 to 0.277) | +0.085 (+0.065 to +0.105) |
| icuDG Transformer | Rockenschaub et al., Crit Care Med 2024 (icuDG) | 0.790 (0.773 to 0.805) | +0.032 (+0.023 to +0.042) | 0.227 (0.210 to 0.247) | +0.065 (+0.050 to +0.080) | 1.05 (0.97 to 1.13) | 0.047 (-0.043 to 0.126) | 0.0207 (0.0173 to 0.0242) | 73.3 (70.6 to 76.1) | 6.3 (6.0 to 6.5) | 0.811 (0.787 to 0.834) | 0.252 (0.229 to 0.278) | +0.084 (+0.065 to +0.103) |
| YAIB Transformer | van de Water et al., ICLR 2024 (YAIB) | 0.788 (0.771 to 0.804) | +0.034 (+0.024 to +0.043) | 0.225 (0.206 to 0.247) | +0.067 (+0.052 to +0.083) | 1.05 (0.98 to 1.13) | 0.056 (-0.033 to 0.133) | 0.0208 (0.0175 to 0.0242) | 74.6 (71.9 to 77.3) | 6.3 (5.9 to 6.5) | 0.812 (0.786 to 0.839) | 0.266 (0.240 to 0.297) | +0.069 (+0.049 to +0.089) |
| NCL | Yeche et al., ICML 2021 (NCL) | 0.788 (0.772 to 0.803) | +0.034 (+0.025 to +0.043) | 0.218 (0.199 to 0.241) | +0.074 (+0.058 to +0.089) | 1.02 (0.95 to 1.09) | 0.042 (-0.049 to 0.121) | 0.0207 (0.0178 to 0.0238) | 74.7 (72.0 to 77.3) | 6.3 (6.0 to 6.6) | 0.803 (0.779 to 0.827) | 0.251 (0.227 to 0.279) | +0.084 (+0.066 to +0.102) |
| icuDG GRU | Rockenschaub et al., Crit Care Med 2024 (icuDG) | 0.787 (0.771 to 0.803) | +0.035 (+0.025 to +0.045) | 0.221 (0.202 to 0.241) | +0.071 (+0.056 to +0.087) | 1.04 (0.97 to 1.12) | 0.046 (-0.048 to 0.128) | 0.0203 (0.0169 to 0.0237) | 74.9 (72.2 to 77.5) | 6.3 (6.0 to 6.6) | 0.812 (0.789 to 0.836) | 0.255 (0.230 to 0.282) | +0.081 (+0.062 to +0.100) |
| SepsisLab LSTM | Yin et al., KDD 2024 (SepsisLab) | 0.787 (0.771 to 0.803) | +0.035 (+0.024 to +0.046) | 0.211 (0.193 to 0.230) | +0.082 (+0.064 to +0.099) | 1.17 (1.09 to 1.27) | 0.070 (-0.020 to 0.152) | 0.0190 (0.0157 to 0.0226) | 73.5 (70.8 to 76.3) | 6.4 (6.0 to 6.7) | 0.807 (0.784 to 0.831) | 0.240 (0.217 to 0.267) | +0.095 (+0.074 to +0.116) |
| Moor attention model | Moor et al., eClinicalMedicine 2023 | 0.786 (0.769 to 0.803) | +0.036 (+0.026 to +0.046) | 0.229 (0.211 to 0.250) | +0.063 (+0.049 to +0.078) | 1.05 (0.98 to 1.12) | 0.066 (-0.022 to 0.143) | 0.0203 (0.0173 to 0.0234) | 74.2 (71.5 to 76.9) | 6.3 (6.0 to 6.6) | 0.803 (0.778 to 0.829) | 0.258 (0.235 to 0.286) | +0.078 (+0.059 to +0.096) |
| YAIB GRU | van de Water et al., ICLR 2024 (YAIB) | 0.784 (0.768 to 0.801) | +0.038 (+0.028 to +0.049) | 0.216 (0.197 to 0.238) | +0.076 (+0.060 to +0.095) | 1.04 (0.96 to 1.13) | 0.057 (-0.034 to 0.135) | 0.0208 (0.0173 to 0.0243) | 74.7 (72.0 to 77.4) | 6.2 (5.8 to 6.5) | 0.811 (0.786 to 0.837) | 0.262 (0.235 to 0.295) | +0.073 (+0.052 to +0.094) |
| GATSM | Kim et al., AAAI 2026 (GATSM) | 0.781 (0.763 to 0.797) | +0.041 (+0.032 to +0.051) | 0.205 (0.189 to 0.222) | +0.087 (+0.070 to +0.104) | 1.05 (0.97 to 1.14) | 0.057 (-0.031 to 0.135) | 0.0200 (0.0167 to 0.0233) | 71.2 (68.5 to 74.0) | 6.5 (6.2 to 6.8) | 0.800 (0.775 to 0.825) | 0.223 (0.205 to 0.245) | +0.113 (+0.091 to +0.134) |
| YAIB TCN | van de Water et al., ICLR 2024 (YAIB) | 0.780 (0.764 to 0.797) | +0.042 (+0.032 to +0.051) | 0.208 (0.191 to 0.229) | +0.084 (+0.068 to +0.100) | 1.07 (0.99 to 1.15) | 0.056 (-0.036 to 0.134) | 0.0196 (0.0163 to 0.0230) | 74.7 (72.0 to 77.5) | 6.2 (5.8 to 6.5) | 0.811 (0.786 to 0.836) | 0.256 (0.230 to 0.287) | +0.079 (+0.059 to +0.099) |
| icuDG TCN | Rockenschaub et al., Crit Care Med 2024 (icuDG) | 0.777 (0.761 to 0.794) | +0.045 (+0.035 to +0.054) | 0.218 (0.200 to 0.238) | +0.074 (+0.059 to +0.090) | 1.00 (0.94 to 1.07) | 0.035 (-0.056 to 0.115) | 0.0188 (0.0157 to 0.0219) | 73.5 (70.7 to 76.3) | 6.2 (5.8 to 6.5) | 0.805 (0.781 to 0.829) | 0.251 (0.229 to 0.279) | +0.084 (+0.067 to +0.102) |
| Tranchellini InceptionTime | Tranchellini et al., npj Digit Med 2026 | 0.775 (0.759 to 0.792) | +0.047 (+0.037 to +0.057) | 0.194 (0.179 to 0.211) | +0.098 (+0.082 to +0.115) | 1.01 (0.94 to 1.09) | 0.040 (-0.051 to 0.120) | 0.0186 (0.0152 to 0.0218) | 73.3 (70.5 to 76.0) | 6.4 (6.1 to 6.7) | 0.799 (0.775 to 0.824) | 0.224 (0.203 to 0.248) | +0.111 (+0.090 to +0.134) |
| Tranchellini CNN | Tranchellini et al., npj Digit Med 2026 | 0.765 (0.748 to 0.781) | +0.057 (+0.047 to +0.068) | 0.185 (0.168 to 0.204) | +0.107 (+0.089 to +0.127) | 1.04 (0.94 to 1.15) | 0.059 (-0.031 to 0.138) | 0.0176 (0.0140 to 0.0212) | 73.7 (70.9 to 76.5) | 6.2 (5.8 to 6.5) | 0.787 (0.761 to 0.814) | 0.229 (0.205 to 0.259) | +0.106 (+0.083 to +0.129) |
| Tranchellini LSTM | Tranchellini et al., npj Digit Med 2026 | 0.747 (0.730 to 0.763) | +0.075 (+0.064 to +0.087) | 0.167 (0.152 to 0.185) | +0.125 (+0.107 to +0.144) | 0.90 (0.82 to 0.98) | 0.031 (-0.061 to 0.111) | 0.0161 (0.0130 to 0.0193) | 75.4 (72.7 to 78.2) | 6.3 (5.9 to 6.5) | 0.772 (0.746 to 0.799) | 0.216 (0.193 to 0.244) | +0.120 (+0.097 to +0.142) |

All 17 models were evaluated on the MIMIC-IV internal test set (4,161 ICU stays, 997 sepsis cases, 182,415 hourly scores; pre-treatment subset: 63,446 scores across 4,160 stays and 996 cases before first in-ICU antimicrobial order and culture). Differences (Δ) are SepsisFlow minus comparator (95% CIs from 2,000 paired patient-cluster bootstrap replicates; Holm-adjusted P = 0.016 for all 32 full-cohort ΔAUROC and ΔAUPRC comparisons). Calibration follows validation Platt scaling; detection uses validation thresholds matched to the NEWS2 ≥ 5 burden (46.0 first alarms per 100 patient-days).